\documentclass[10pt,twocolumn,letterpaper]{article}

\usepackage[pagenumbers]{iccv} 
\usepackage{graphicx}
\usepackage{booktabs} 
\usepackage{microtype}

\usepackage{algorithmic}
\usepackage{algorithm}
\usepackage{amsmath}
\usepackage{amssymb}
\usepackage{mathtools}
\usepackage{amsthm}
\usepackage{booktabs}
\definecolor{darkblue}{rgb}{0, 0, 0.5}
\usepackage{wrapfig}
\usepackage{xspace}
\usepackage{lipsum}
\usepackage{threeparttable}
\usepackage{tablefootnote}
\usepackage{colortbl}
\usepackage{array, arydshln} 

\definecolor{iccvblue}{rgb}{0.21,0.49,0.74}
\usepackage[pagebackref,breaklinks,colorlinks,allcolors=iccvblue]{hyperref}

\def\confName{ICCV}
\def\confYear{2025}

\title{Mol-CADiff: Causality-Aware Autoregressive Diffusion for Molecule Generation}

\author{
Md Atik Ahamed$^{1}$, Qiang Ye$^{2}$, Qiang Cheng$^{1,3}$\thanks{Corresponding author.} \\
$^{1}$Department of Computer Science\\
$^{2}$Department of Mathematics\\
$^{3}$Institute for Biomedical Informatics\\
University of Kentucky\\
{\tt\small \{atikahamed,qye3,qiang.cheng\}@uky.edu}
}

\begin{document}
\maketitle
\begin{abstract}
The design of novel molecules with desired properties is a key challenge in drug discovery and materials science. Traditional methods rely on trial-and-error, while recent deep learning approaches have accelerated molecular generation. However, existing models struggle with generating molecules based on specific textual descriptions. We introduce Mol-CADiff, a novel diffusion-based framework that uses causal attention mechanisms for text-conditional molecular generation. Our approach explicitly models the causal relationship between textual prompts and molecular structures, overcoming key limitations in existing methods. We enhance dependency modeling both within and across modalities, enabling precise control over the generation process. Our extensive experiments demonstrate that Mol-CADiff outperforms state-of-the-art methods in generating diverse, novel, and chemically valid molecules, with better alignment to specified properties, enabling more intuitive language-driven molecular design.
\end{abstract}
\section{Introduction}
The design and discovery of novel molecules with desired properties is a foundational challenge in drug discovery, materials science, and chemical engineering. Traditional approaches to molecular design often rely on trial-and-error experimentation or computational methods that explore only limited regions of the vast chemical space. Recent advances in deep learning have shown promising results in automating and accelerating this process, enabling the generation of molecules with specified characteristics.

Molecular generation presents unique challenges that distinguish it from other generative tasks. Molecules must adhere to physical and chemical constraints to be valid and synthesizable. Moreover, the discrete nature of molecular structures, represented as graphs with atoms as nodes and bonds as edges, adds complexity compared to continuous domains like images or audio. Effective molecular generation requires models that can capture both local atomic interactions and global molecular properties while maintaining chemical validity.

Recent work has leveraged various deep learning architectures for this task, including variational autoencoders (VAEs), autoregressive models, and generative adversarial networks (GANs). While these approaches have shown success in generating valid molecules, they often struggle with controlled generation according to specific textual descriptions or property requirements. This limitation has motivated research into multimodal approaches that can bridge the gap between natural language descriptions and molecular structures.

Diffusion models have emerged as a powerful paradigm for generative modeling, achieving state-of-the-art (SOTA) results across various domains including images, audio, and 3D structures. Their application to molecular generation has shown promise, particularly in unconditional settings. However, existing diffusion-based approaches for controlled molecular generation often rely on simple architectures that fail to fully capture the complex dependencies between textual descriptions and molecular structures.

In this paper, we introduce Mol-CADiff, a novel causal attention diffusion framework for text-conditional molecular generation. Our approach addresses key limitations of existing methods by explicitly modeling the causal relationship between textual prompts and molecular structures, enabling more precise control over the generation process. Unlike previous approaches that rely on simple MLP-based denoising networks, Mol-CADiff employs a sophisticated attention mechanism that enhances dependency modeling both within and across modalities.

Our contributions can be summarized as follows:  
\begin{itemize}  
\item We develop an attention-based denoising module for the diffusion network, enhancing dependency modeling both across modalities and within each modality. This contrasts with SOTA models like 3M-Diffusion, which rely on MLP-based denoising.  
\item We integrate partial tokens from the clean graph latent representation alongside conditional and noisy latent inputs, enriching contextual information for more effective molecule generation.  
\item Our approach explicitly models the causal relationship between text prompts and molecular graphs, ensuring structured generation. To our knowledge, this is the first diffusion-based molecular generation model to incorporate causal dependencies in an autoregressive fashion. 
\item We conduct extensive experiments, demonstrating the superiority of our method over SOTA baselines in generating diverse, novel, and chemically valid molecules.  
\end{itemize}  

Our comprehensive evaluation on standard molecular generation benchmarks demonstrates that Mol-CADiff outperforms existing methods across multiple metrics, including validity, novelty, uniqueness, and alignment with specified properties. Furthermore, we show that our approach enables fine-grained control over molecular generation through natural language instructions, opening new possibilities for intuitive and accessible molecular design by domain experts without extensive programming knowledge.

\section{Related Work}
{\bf{Molecular Representation and Generation}}.
Early approaches to molecular generation relied on string-based representations such as SMILES \citep{weininger1988smiles} and SELFIES \citep{krenn2022selfies}, which encode molecular structures as linear strings. While these representations are compact, they often struggle with maintaining chemical validity in generation tasks \citep{bjerrum2017molecular, gomez2018automatic}. More recent work has explored graph-based approaches that explicitly model molecular structures as graphs, where atoms are represented as nodes and bonds as edges \citep{li2018learning}.

{\bf{Graph-Based Molecular Generation}}. 
Graph-based methods have demonstrated significant advantages in generating valid, diverse, and novel molecular structures. Early graph-based approaches include variational autoencoders (VAEs) \citep{liu2018constrained, jin2018junction}, autoregressive models \citep{you2018graph, liao2019efficient, popova2019molecularrnn, goyal2020graphgen}, and flow-based models \citep{madhawa2019graphnvp, zang2020moflow, luo2021graphdf}. These methods have progressively improved in generating molecules with desired properties and structural constraints.
A parallel line of research has focused on fragment-based approaches \citep{jin2020hierarchical, kong2022molecule}, which decompose molecules into substructures and then recombine them, aligning with the fragment-based drug design paradigm in medicinal chemistry \citep{hajduk2007decade}. These methods have shown promise in maintaining chemical validity while exploring diverse chemical spaces.

{\bf{Diffusion Models for Molecular Generation}}.
Recent advances in diffusion models \citep{ho2020denoising, song2020denoising} have spurred their application to molecular generation \citep{vignac2022digress, jo2022score}. Diffusion models have shown remarkable capabilities in generating high-quality samples across various domains \citep{rombach2022high}. In the molecular domain, these models offer a balance between exploration and validity, addressing limitations of previous approaches.

{\bf{Language Models and Molecular Generation}}.
The success of large language models (LLMs) in natural language processing has inspired their application to molecular design \citep{edwards2022translation, flam2022language, fang2023molecular}. Models like MolT5 \citep{edwards2022translation} and ChemT5 \citep{christofidellis2023unifying} adapt the T5 architecture \citep{raffel2020exploring} for molecular tasks, treating molecules as a specialized form of language. Mol-Instructions \citep{fang2023mol} further enhances these capabilities by incorporating instruction-following abilities, enabling more directed molecular generation.

{\bf{Multimodal Approaches}}. 
Most recently, multimodal approaches that bridge textual descriptions and molecular structures have emerged \citep{edwards2021text2mol, zeng2022deep, su2022molecular, MolBind2024}. These models aim to leverage the complementary strengths of language and molecular representations, enabling more intuitive and flexible control over generation through natural language instructions. 3M-Diffusion \citep{zhu20243m} represents a significant advance in this direction by employing a latent diffusion approach that bridges molecular and textual modalities.

Our work, Mol-CADiff, addresses key limitations in previous approaches. Unlike methods that treat text-to-molecule generation as simple translation tasks without causal modeling, we combine autoregression with diffusion models using a novel cross-modal attention mechanism. We replace MLP-based denoising with an attention architecture that enhances dependency modeling across modalities, while explicitly modeling causal relationships between text prompts and molecular graphs. This approach enables more precise control over molecular properties while maintaining coherence with textual descriptions, advancing the state-of-the-art in conditional molecular generation.

\section{Methodology}

In this section, we present our method step by step. We consider a molecule represented as a graph $G$ along with its corresponding textual description $C$ as a condition. Our goal is to develop a causality-aware auto-regressive framework that generates novel, diverse, and chemically valid molecules while preserving the properties described in the given condition. Formally, a molecule is represented as $G = (V, A)$, where $V$ is the set of $n$ nodes, and $A \in \mathbb{R}^{n \times n \times a}$ is the adjacency matrix encoding edge features. Each entry in $A$ is one-hot encoded, where each distinct value corresponds to a specific bond type.
\begin{figure*}
    \centering
    \includegraphics[width=0.9\linewidth]{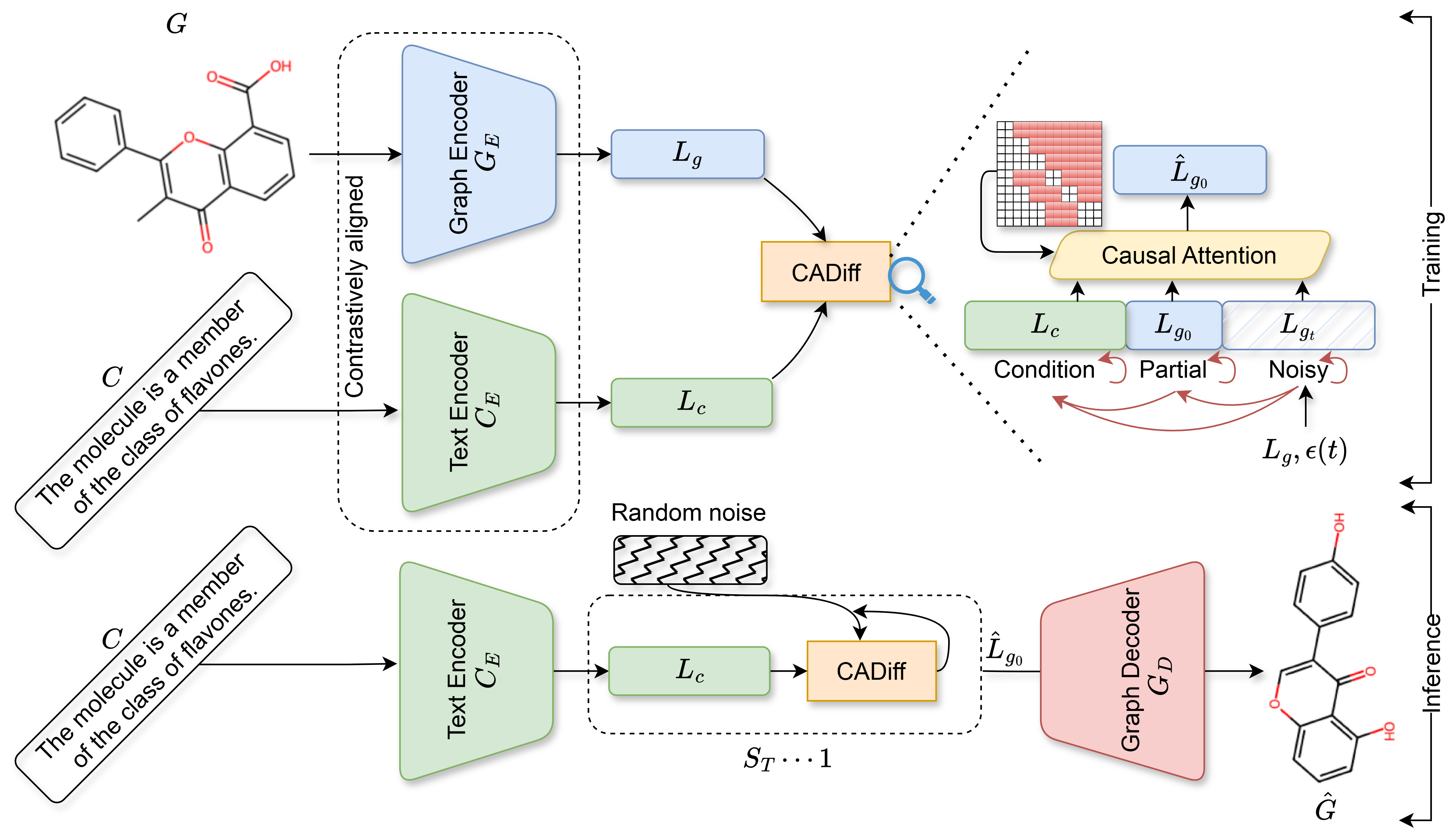}
    \caption{Schematic of Mol-CADiff: Encoders map graphs and text to latent spaces ($G \to L_g$, $C \to L_c$). CADiff integrates autoregression with diffusion, ensuring causal dependencies and multimodal attention. During inference, instruction $C$ guides iterative denoising, producing $\hat{L}_{g_0}$, which $G_D$ decodes into a molecule.}

    \label{fig:method}
\end{figure*}
\subsection{Pretraining Stages}  
To enable efficient diffusion modeling, we first transform molecular graphs and their corresponding textual descriptions into a compact latent space. This reduces computational complexity by operating on fewer tokens rather than the full graph-text representation. We employ two pretraining strategies: contrastive learning for aligning graph and text representations and encoder-decoder training for structured graph reconstruction.

First, we utilize a pretrained Graph Encoder $G_E$ and Text Encoder $C_E$, both optimized using contrastive learning, for example GIN~\cite{hu2019strategies} as $G_E$ 
 and  Sci-BERT~\cite{beltagy2019scibert} as $C_E$. The latent representations, $L_g$ and $L_c$, extracted from $G_E$ and $C_E$, respectively, are aligned via contrastive learning on a dataset of molecule-text pairs~\citep{liu2023molca}, enforcing similarity between corresponding pairs while distinguishing unrelated ones.

Next, we train the Graph Encoder-Decoder pair ($G_E$, $G_D$) in an autoencoder-style manner to reconstruct molecular structures. The Graph Decoder $G_D$ (e.g., HierVAE~\cite{jin2020hierarchical}) reconstructs the molecular graph $\hat{G}$ from the latent representation $L_g$, optimizing the negative Evidence Lower Bound (ELBO)~\citep{zhu20243m}. These pretraining steps reduce the complexity of diffusion modeling by operating in a structured latent space rather than raw molecular graphs and text, thereby improving computational efficiency.

\subsection{Integrating Autoregression with Diffusion}
To alleviate diffusion modeling and generate diverse, novel, and chemically valid molecules, we introduce an autoregressive causal dependency as demonstrated in Figure~\ref{fig:method} under the CADiff module. This module requires three key latent representations: the conditional text-based latent representation $L_c$, the partial clean latent representation of the molecular graph $L_{g_0}$, and the noisy graph latent representation $L_{g_t}$ at diffusion timestep $t$. For consistency, we denote the unperturbed (clean) latent representation $L_g$ as $L_{g_0}$, corresponding to diffusion timestep $t=0$. The forward diffusion process progressively perturbs $L_{g_0}$ by adding Gaussian noise, producing a sequence of noisy latents $\{L_{g_1}, L_{g_2}, ..., L_{g_T}\}$. This follows a Markovian diffusion process, where at each step $t$, the latent representation is perturbed based on a predefined variance schedule:
\begin{equation}
    q(L_{g_t} | L_{g_{t-1}}) = \mathcal{N} \left( L_{g_t} ; \sqrt{1 - \beta_t} L_{g_{t-1}}, \beta_t I \right),
\end{equation}
where $\beta_t$ is the variance schedule at timestep $t$, and $I$ is the identity matrix. The variance schedule $\{\beta_t\}_{t=1}^{T} $ is a monotonically increasing sequence, ensuring a gradual injection of noise. The full forward diffusion process, allowing direct computation of $L_{g_t}$ from $L_{g_0}$, is given by:
\begin{equation}
    q(L_{g_t}|L_{g_0}) = \mathcal{N} \left (L_{g_t};\sqrt{\bar{\alpha}_t} L_{g_0}, (1 - \bar{\alpha}_t) I\right),
\end{equation}
\begin{equation}
    L_{g_t} = \sqrt{\bar{\alpha}_t} L_{g_0} + \sqrt{1 - \bar{\alpha}_t} \epsilon, \quad \text{where } \epsilon \sim \mathcal{N}(0, I),
\end{equation}
where $\bar{\alpha}_t = \prod_{s=1}^{t} (1 - \beta_s)$ is the cumulative product of noise reduction terms. The full joint probability distribution over all diffusion steps is then factorized as:
\begin{equation}
    q(L_{g_{0:T}}) = q(L_{g_0}) \prod_{t=1}^{T} q(L_{g_t} | L_{g_{t-1}}).
\end{equation}
This formulation effectively models the progressive noise injection process while preserving structural dependencies essential for molecular generation. While diffusion models are well suited for image generation and denoising tasks, they inherently lack the ability to capture dependencies in a causal autoregressive manner.

To address this, we incorporate causal attention in place of regular attention and adopt an autoregressive token generation strategy, effectively integrating autoregression with diffusion. This autoregressive causality is enforced through a structured causal attention mechanism, where conditional tokens, partial clean graph tokens, and noisy tokens interact within the denoising process. By incorporating both clean and noisy graph tokens alongside conditional tokens, our approach ensures effective multimodal integration. To maintain dependencies across tokens, we partition them into AR steps and introduce a generalized causal attention mask, as detailed in Subsection~\ref{sub:method_ar_step}.

Unlike image generation, which relies on spatial dependencies within patches and class labels, molecular graph data involves high-dimensional continuous values and complex relational structures. Our approach effectively addresses these challenges through four key components: \textbf{\textit{(1)}} a conditional causality-aware attention mechanism that captures continuous expression levels from text embeddings rather than discrete class labels; \textbf{\textit{(2)}} a token-based processing system, where compressed graph and text representations function as tokens; \textbf{\textit{(3)}} preservation of token ordering across multiple modalities rather than shuffling them; and \textbf{\textit{(4)}} a denoising strategy that directly predicts the clean version of tokens instead of modeling added noise, diverging from CausalFusion~\citep{deng2024causal}.

This integration of autoregression and diffusion enables the model to capture structural dependencies while effectively generating chemically valid, novel, and diverse molecular graphs. More formally, the integration of AR with diffusion can be represented by,
\begin{equation}
\label{eq:ar_fact}
\resizebox{\linewidth}{!}{$
q({L_{g_{0:T,\kappa_s}}} | {L_{g_{0,\kappa_{1:s-1}}}}) = q(L_{g_{0,\kappa_s}})  
\prod_{t=1}^{T} q(L_{g_{t,\kappa_s}} | L_{g_{t-1,\kappa_s}}, L_{g_{0,\kappa_{1:s-1}}}).
$}
\end{equation}
In Equation~\ref{eq:ar_fact}, $S$ represents the number of AR steps, where $s \in [1, S]$, and $\kappa_s$ denotes the index set of tokens processed at the $s^{th}$ AR step. The number of tokens in each AR step is given by $|\kappa_s|$, and only the tokens indexed by $\kappa_s$ are processed during that step. Here, $L_{g_{t,\kappa_s}}$ represents the dual-factorized graph tokens at the $s^{th}$ AR step and $t^{th}$ diffusion step. During training, our objective is to approximate $p_\theta(L_{g_{t-1},\kappa_s} | L_{g_{t,\kappa_s}},L_{g_{0,\kappa_{1:s-1}}})$ for all $t$ and $s$. This estimation also requires the partial clean latent representation $L_{g_{0,\kappa_{1:s-1}}}$ with the noisy tokens of the current AR step.

This structure allows the model to leverage information from previous AR steps, improving the denoising of the current tokens. A generalized causal attention mask ensures that $L_{g_{0,\kappa_{1:s-1}}}$ does not have access to $L_{g_{t,\kappa_s}}$, enforcing proper sequential dependencies. We illustrate an example of this causal attention mechanism in Figure~\ref{fig:method} and Figure~\ref{fig:full_part}. Effectively, this can be interpreted as a next-token prediction task, where each token attends to previously generated tokens while simultaneously being denoised through diffusion. By incorporating the causal attention mask, we maintain structured dependencies both within the same modality (intra-modal) and across different modalities (inter-modal).

\subsection{AR Steps Formulation}
\label{sub:method_ar_step}
We generate AR steps ($as$) using Algorithm~\ref{alg:ar_steps}, which takes the sample length $l$ and decay factor $\gamma$ as inputs. When $\gamma = 1.0$, the AR step count is uniformly sampled, leading to an unbiased distribution of step sizes. For $\gamma < 1.0$, smaller step sizes are favored at the beginning due to the exponential decay, resulting in progressively larger steps later in the sequence. This behavior affects how information is structured across autoregressive steps, and we further analyze its effect in the ablation study (Subsection~\ref{sub:decay}).
\begin{algorithm}[H]
\caption{Generate AR Steps}
\label{alg:ar_steps}
\textbf{Input:} $l$ (length), $\gamma$ (decay)
\begin{algorithmic}[1]
\STATE $I \gets \text{Uniform}([1, l])$ if $\gamma = 1.0$, else sample from $[1, l]$ with $p_i = b \cdot \gamma^i$ for $i\in[0, l-1]$ , where $b = \frac{1 - \gamma}{1 - \gamma^l}$
\STATE $cm \gets [0] \cup \text{sort}(\text{random sample from } [1, l) \text{ of size } I-1) \cup [l]$
\STATE $as \gets [cm[i+1] - cm[i] \mid i \in [0, I-1]]$
\end{algorithmic}
\textbf{return} $as, cm$
\end{algorithm}

\subsection{Inferencing from instruction}
During the inference (test) phase, molecular generation depends on whether the task is conditional or unconditional. For conditional generation, we use only the provided instruction $C$ to guide the process, whereas for unconditional generation, no such instruction is required. While Figure~\ref{fig:method} illustrates the framework for conditional generation, our method seamlessly extends to unconditional generation by omitting the conditioning information from $L_c$. During inference, we initialize the process with random noise and iteratively refine it over sampling timesteps, from $S_T$ down to 1. The CADiff module is responsible for this iterative denoising, progressively reconstructing the latent representation. Once the denoised latent representation $\hat{L}_{g_0}$ is obtained, it is passed through the Graph Decoder $G_D$, which generates the final molecular structure $\hat{G}$. This approach ensures a structured, stepwise refinement of molecular representations, leading to high-quality, valid, and diverse molecular samples.  

\section{Experiments}
\begin{table*}[!ht]
\centering
\caption{Quantitative comparison of conditional generation on all four datasets. Mol-CADiff outperforms other SOTA methods by a large margin in novelty (Nov.), diversity (Div.), and validity (Val.) metrics while maintaining strong similarity (Sim.) and consistently outperforming on Overall (Ove.) metric. Results are presented in percentage \% values (higher indicating better).}
\label{tab:cond_all}
\resizebox{\textwidth}{!}{
\begin{threeparttable}
\begin{tabular}{l | c c c c c | c c c c c | c c c c c | c c c c c}
    \toprule
    & \multicolumn{5}{c|}{\textbf{ChEBI-20}} & \multicolumn{5}{c|}{\textbf{PubChem}} & \multicolumn{5}{c|}{\textbf{PCDes}} & \multicolumn{5}{c}{\textbf{MoMu}} \\
    Methods & Sim.& Nov.& Div.& Val.& Ove.& Sim.& Nov.& Div.& Val.& Ove.& Sim.& Nov.& Div.& Val.& Ove.& Sim. & Nov.& Div.& Val.& Ove.\\
    \midrule
    MolT5-small & 73.32 & 31.43 & 17.22 & 78.27 & 50.06 & 68.36 & 20.63 & 9.32 & 78.86 & 44.29 & 64.84 & 24.91 & 9.67 & 73.96 & 43.35 & 16.64 & 97.49 & 29.95 & 60.19 & 51.07 \\
    MolT5-base  & 80.75 & 32.83 & 17.66 & 84.63 & 53.97 & 73.85 & 21.86 & 9.89 & 79.88 & 46.37 & 71.71 & 25.85 & 10.50 & 81.92 & 47.50 & 19.76 & 97.78 & 29.98 & 68.84 & 54.09 \\
    MolT5-large & 96.88 & 12.92 & 11.20 & 98.06 & 54.77 & 91.57 & 20.85 & 9.84 & 95.18 & 54.36 & 88.37 & 20.15 & 9.49 & 96.48 & 53.62 & 25.07 & 97.47 & 30.33 & 90.40 & 60.82 \\
    ChemT5-small & 96.22 & 13.94 & 13.50 & 96.74 & 55.10 & 89.32 & 20.89 & 13.10 & 93.47 & 54.19 & 86.27 & 23.28 & 13.17 & 93.73 & 54.11 & 23.25 & 96.97 & 30.04 & 88.45 & 59.68 \\
    ChemT5-base  & 95.48 & 15.12 & 13.91 & 97.15 & 55.42 & 89.42 & 22.40 & 13.98 & 92.43 & 54.56 & 85.01 & 25.55 & 14.08 & 92.93 & 54.39 & 23.40 & 97.65 & 30.07 & 87.61 & 59.68 \\
    Mol-Instruction & 65.75 & 32.01 & 26.50 & 77.91 & 50.54 & 23.40 & 37.37 & 27.97 & 71.10 & 39.96 & 60.86 & 35.60 & 24.57 & 79.19 & 50.06 & 14.89 & 97.52 & 30.17 & 68.32 & 52.73 \\
    3M-Diffusion & 87.09 & 55.36 & 34.03 & 100.0 & 69.12 & 87.05 & 64.41 & 33.44 & 100.0 & 71.22 & 81.57 & 63.66 & 32.39 & 100.0 & 69.41 & 24.62 & 98.16 & 37.65 & 100.0 & 65.11 \\
    \midrule
    \rowcolor{blue!10} \textbf{Mol-CADiff} & 81.92 & 67.35 & 75.69 & 100.0 & \textbf{81.24} & 78.35 & 71.97 & 75.39 & 100.0 & \textbf{81.43} & 71.41 & 74.77 & 75.45 & 100.0 & \textbf{80.41} & 24.41 & 98.34 & 83.30 & 100.0 & \textbf{76.51} \\
    \bottomrule
\end{tabular}
\end{threeparttable}
}
\end{table*}
\begin{table}[t]
\centering
\caption{Quantitative comparison of unconditional generation. Results of Uniq (Uni.), KL Div (KL), and FCD on ChEBI-20 and PubChem, which refer to Uniqueness, KL Divergence, and Fréchet ChemNet Distance, respectively. Results are presented in percentage values. A higher number indicates a better generation quality.}
\label{tab:uncond}
\resizebox{\linewidth}{!}{
\begin{threeparttable}
\begin{tabular}{l | c c c c c c | c c c c c c}
    \toprule
    & \multicolumn{6}{c|}{\shortstack[c]{\textbf{ChEBI-20}}} & \multicolumn{6}{c}{\shortstack[c]{\textbf{PubChem}}} \\
    Methods & Uni.& Nov. &  KL & FCD& Val. & Ove. & Uni.& Nov.&  KL& FCD& Val. & Ove.\\
    \midrule
    CharRNN & 72.46 & 11.57 & 95.21 & 75.95  & 98.21 & 70.68 & 63.28 & 23.47 & 90.72 & 76.02 & 94.09 & 69.52 \\
    VAE & 57.57 & 47.88 & 95.47 & 74.19  & 63.84& 67.79 & 44.45& 42.47 & 91.67 & 55.56 & 94.10 &65.65 \\
    AAE & 1.23 & 1.23 & 38.47 & 0.06  & 1.35 & 8.47 & 2.94 & 3.21 & 39.33 & 0.08 & 1.97  & 9.51\\
    LatentGAN & 66.93 & 57.52 & 94.38 & 76.65  & 73.02& 73.70& 52.00 & 50.36 & 91.38 & 57.38 & 53.62 & 60.95 \\
    BwR & 22.09 & 21.97 & 50.59 & 0.26 & 22.66 & 23.51& 82.35 & 82.34 & 45.53 & 0.11 & 87.73 & 59.61\\ 
    HierVAE & 82.17 & 72.83 & 93.39 & 64.32 & 100.0 & 82.54& 75.33 & 72.44 & 89.05 & 50.04 & 100.0  &77.37\\ 
    PS-VAE & 76.09 & 74.55 & 83.16 & 32.44 & 100.0& 73.25& 66.97 & 66.52 & 83.41 & 14.41 & 100.0  & 66.26\\
    \textbf{3M-Diffusion} & 83.04 & 70.80 & 96.29 & 77.83 & 100.0& 85.59& 85.42 & 81.20 & 92.67 & 58.27 & 100.0 & 83.51\\
    \midrule
    \rowcolor{blue!10} \textbf{Mol-CADiff}& 85.90 & 79.11 & 94.69 & 74.45 &100.0 & \bf86.83& 87.79 & 83.23 & 95.19 & 71.65 & 100.0 &\bf87.57\\
    \bottomrule
\end{tabular}
\end{threeparttable}
}
\end{table}
In this section, we outline the experimental setup and provide insights into the obtained results.

\paragraph{Datasets:} We utilize four molecular datasets: PubChem~\citep{liu2023molca}, ChEBI-20~\citep{edwards2021text2mol}, PCDes~\citep{zeng2022deep}, and Momu~\citep{su2022molecular}. To ensure a fair comparison, we follow prior studies on molecular structure generation~\citep{irwin2012zinc, blum2009970, rupp2012fast, zhu20243m} and restrict the dataset to molecules with fewer than 30 atoms, aligning with the approach in~\citet{zhu20243m}. Table~\ref{tab:stats} presents the dataset statistics. PCDes and MoMu utilize the same training and validation set while maintaining different test sets. We present and compare the obtained results on the test set.
\begin{table}[b]
    \centering
    \caption{Dataset statistics for training, validation, and testing.}
    \begin{tabular}{lccc}
        \toprule
        \textbf{Dataset} & \textbf{Training} & \textbf{Validation} & \textbf{Test} \\
        \midrule
        ChEBI-20  & 15,409  & 1,971  & 1,965  \\
        PubChem   & 6,912   & 571    & 1,162  \\
        PCDes     & 7,474   & 1,051  & 2,136  \\
        MoMu      & 7,474   & 1,051  & 4,554  \\
        \bottomrule
    \end{tabular}
    \label{tab:stats}
\end{table}
\paragraph{Metrics:} 
We evaluate model performance on text-guided conditional and unconditional molecule generation. For text-guided generation, we follow prior works~\citep{edwards2022translation, christofidellis2023unifying, fang2023mol,zhu20243m} and assess (\textbf{i}) \textit{Similarity}, the percentage of generated molecules matching the ground truth with MACCS~\citep{durant2002reoptimization} cosine similarity with a threshold of $0.5$; (\textbf{ii}) \textit{Novelty}, the fraction of qualified molecules with $f(G, \hat{G})< 0.8$; (\textbf{iii}) \textit{Diversity}, the average pairwise distance $1 - f(\cdot, \cdot)$ between qualified molecules; and (\textbf{iv}) \textit{Validity}, the percentage of chemically valid molecules.

For unconditional generation, we use the GuacaMol benchmarks~\citep{brown2019guacamol} and measure (\textbf{i}) \textit{Uniqueness}, the ratio of distinct molecules; (\textbf{ii}) \textit{Novelty}, the fraction not seen in training; (\textbf{iii}) \textit{KL Divergence}, which quantifies alignment with training set distributions; and (\textbf{iv}) \textit{Fréchet ChemNet Distance} (FCD)~\citep{preuer2018frechet}, which compares feature distributions in ChemNet. All metrics range from 0 to 1 and finally showed in percentage value (out of 100\%), with higher values indicating better performance. However, as observed in Tables~\ref{tab:cond_all} and \ref{tab:uncond}, some models excel in specific metrics (e.g., MolT5-large on Similarity) but underperform in others (e.g., lower Novelty and Diversity). To provide a holistic comparison, we introduce an (\textbf{v}) \textit{Overall} metric, which computes the average of all individual metrics. This offers a balanced assessment of a model's general performance rather than emphasizing only one aspect.

\subsection{Implementation Details}
Here, we outline our implementation for conditional and unconditional molecule generation and briefly introduce the baseline models for comparison.

\paragraph{Baselines:} 
For conditional/instruction-guided molecule generation, we evaluate Mol-CADiff against four baseline models: MolT5~\citep{edwards2022translation}, ChemT5~\citep{christofidellis2023unifying}, Mol-Instruction~\citep{fang2023mol} and 3M-Diffusion~\cite{zhu20243m} with variations from MolT5 (Small, base and large) and ChemT5 (small and base).

For unconditional molecule generation, we compare against eight recent Variational Autoencoder (VAE)-based methods: CharRNN~\citep{segler2018generating}, VAE~\citep{kingma2013auto}, AAE~\citep{makhzani2015adversarial}, LatentGAN~\citep{prykhodko2019novo}, BwR~\citep{diamant2023improving}, HierVAE~\citep{jin2020hierarchical}, and PS-VAE~\citep{kong2022molecule} and 3M-Diffusion~\cite{zhu20243m}. We followed the key experimental protocols from 3M-Diffusion including the training, validation, and test set splits, and presented the baseline result from the paper to ensure fair comparison and to avoid any potential under-performing results from the baselines.
\begin{figure*}[t]
    \centering
    \includegraphics[width=0.7\linewidth,  trim={0cm 0.8cm 0cm 0cm}, clip]{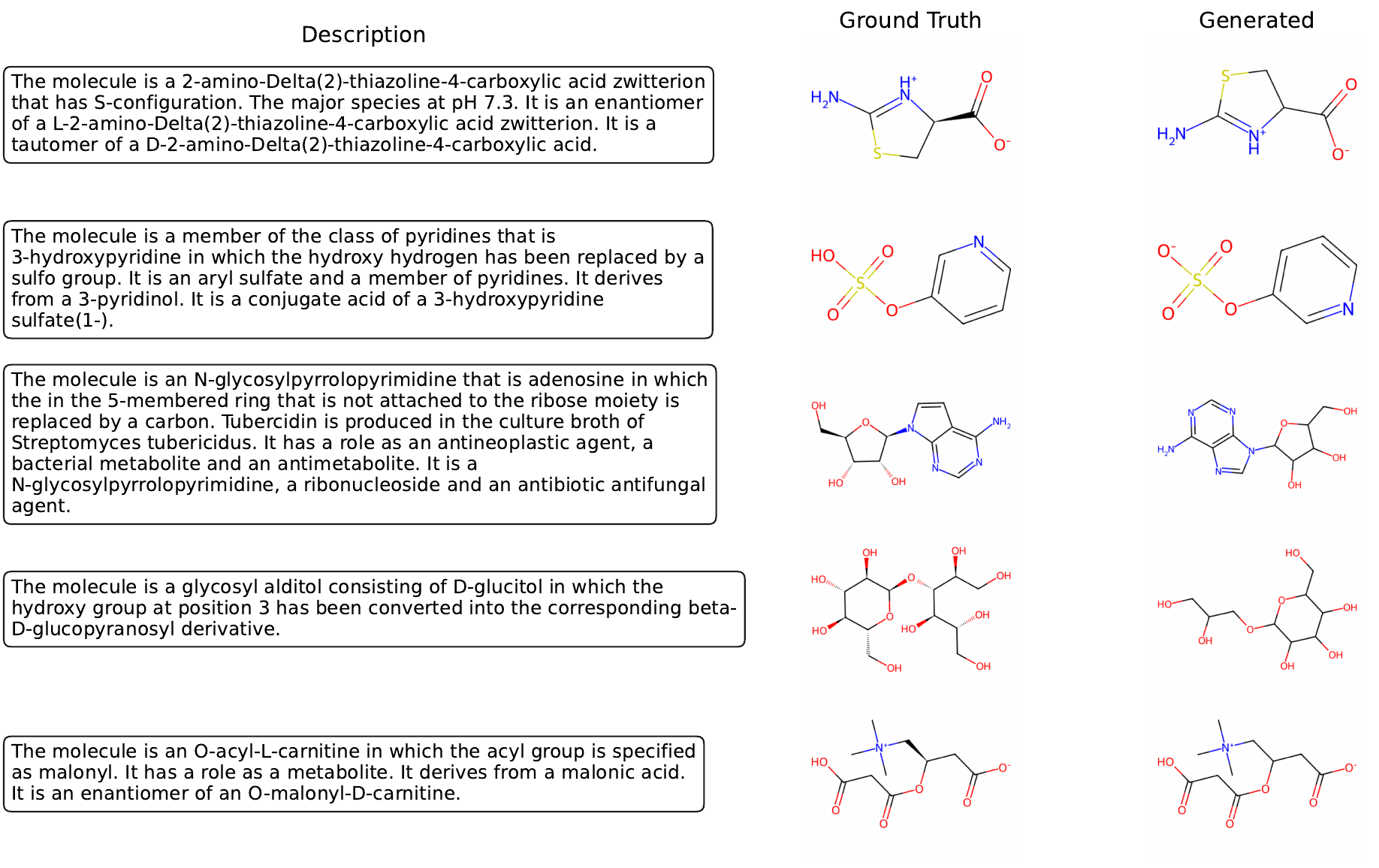}
    \caption{Conditional molecule generation based on text prompt from ChEBI-20 test set (unseen) demonstrating the generated molecules retaining the textual information and maintaining similarity with ground truth (best viewed when zoomed in).}
    \label{fig:qual}
\end{figure*}
\begin{table*}[!ht]
    \centering
    \resizebox{0.85\textwidth}{!}{
    \begin{tabular}{lccccc}
        & \multicolumn{5}{c}{\textbf{Condition:} This molecule is insoluble in water.}\\
         \rotatebox{90}{Mol-CADiff}&  \includegraphics[width=0.15\linewidth,trim={0cm 0.8cm 0cm 1.1cm}, clip]{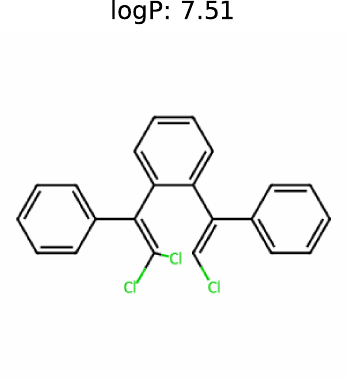}&  \includegraphics[width=0.15\linewidth,trim={0cm 0.8cm 0cm 1.1cm}, clip]{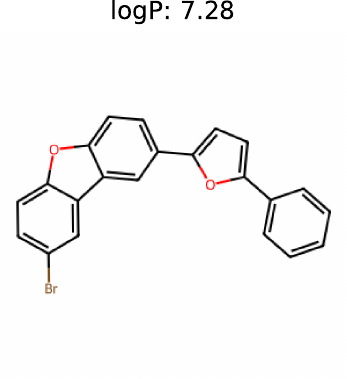}&  \includegraphics[width=0.15\linewidth,trim={0cm 1cm 0cm 1.1cm}, clip]{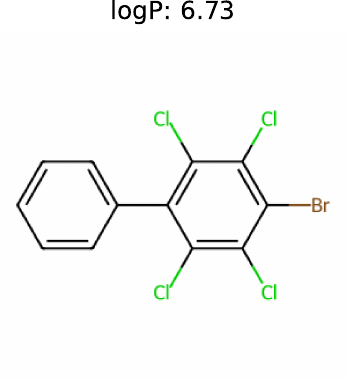}&  \includegraphics[width=0.15\linewidth,trim={0cm 0.8cm 0cm 1.1cm}, clip]{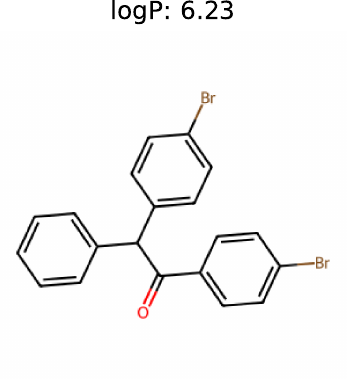}&  \includegraphics[width=0.15\linewidth,trim={0cm 0.8cm 0cm 1.1cm}, clip]{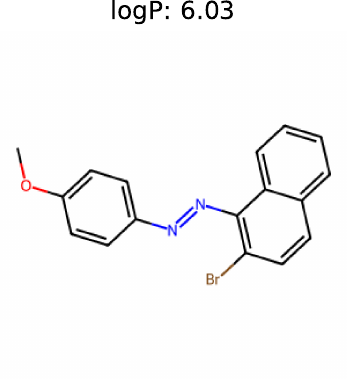}\\
         & logP: 7.51 & logP: 7.28 & logP: 6.73 & logP: 6.23 & logP: 6.03\\
        \rotatebox{90}{MolT5-large}&  \includegraphics[width=0.15\linewidth,trim={9.3cm 35.4cm 45cm 2.9cm}, clip]{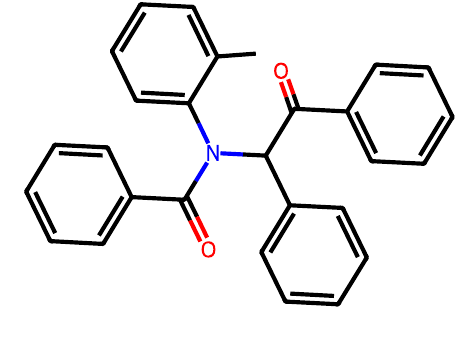} & \includegraphics[width=0.15\linewidth,trim={20.3cm 35.4cm 34cm 2.9cm}, clip]{figures/case_study.pdf} & \includegraphics[width=0.15\linewidth,trim={31.3cm 35.4cm 23cm 2.9cm}, clip]{figures/case_study.pdf} & \includegraphics[width=0.15\linewidth,trim={43.3cm 35.4cm 11cm 2.9cm}, clip]{figures/case_study.pdf} & \includegraphics[width=0.15\linewidth,trim={53.3cm 35.4cm 1cm 2.9cm}, clip]{figures/case_study.pdf}\\
        & logP: 5.67 & logP: 5.67 & logP: 5.67 & logP: 5.67 & logP: 1.32\\
        \rotatebox{90}{3M-Diffusion}&  \includegraphics[width=0.15\linewidth,trim={7.57cm 28cm 45cm 13.1cm}, clip]{figures/case_study.pdf} &  \includegraphics[width=0.15\linewidth,trim={18.56cm 26cm 35cm 13cm}, clip]{figures/case_study.pdf} &  \includegraphics[width=0.15\linewidth,trim={30.56cm 26cm 23.4cm 13cm}, clip]{figures/case_study.pdf} &  \includegraphics[width=0.15\linewidth,trim={41.56cm 26cm 11.4cm 12cm}, clip]{figures/case_study.pdf} &  \includegraphics[width=0.15\linewidth,trim={52.56cm 26cm 0.4cm 12cm}, clip]{figures/case_study.pdf}\\
        & logP: 6.59 & logP: 6.22 & logP: 6.03 & logP: 5.81 & logP: 5.67\\
        \cdashline{2-6}
         & \multicolumn{5}{c}{\textbf{Condition:} The molecule is a member of the class of flavones.}\\
         \rotatebox{90}{Mol-CADiff}&  \includegraphics[width=0.11\linewidth,trim={0cm 1cm 0cm 1.1cm}, clip]{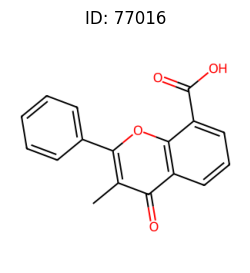}&  \includegraphics[width=0.15\linewidth,trim={0cm 1.7cm 0cm 2.2cm}, clip]{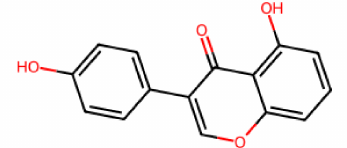}&  \includegraphics[width=0.15\linewidth,trim={0cm 1.7cm 0cm 2.2cm}, clip]{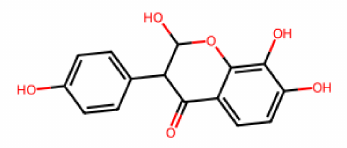}&  \includegraphics[width=0.15\linewidth,trim={0cm 1.5cm 0cm 2cm}, clip]{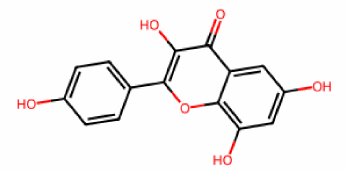}&  \includegraphics[width=0.15\linewidth,trim={0cm 0.8cm 0cm 1.1cm}, clip]{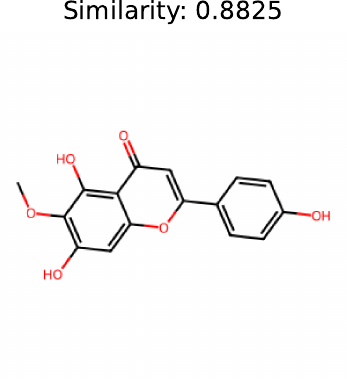}\\
         & Reference & Sim: 0.93 & Sim: 0.91 & Sim: 0.89 & Sim: 0.88\\
        \rotatebox{90}{MolT5-large}&  \includegraphics[width=0.11\linewidth,trim={0cm 1cm 0cm 1.1cm}, clip]{figures/reference.png} & \includegraphics[width=0.15\linewidth,trim={19.3cm 15cm 34.8cm 23.9cm}, clip]{figures/case_study.pdf} & \includegraphics[width=0.15\linewidth,trim={29.3cm 15.4cm 22cm 23.9cm}, clip]{figures/case_study.pdf} & \includegraphics[width=0.15\linewidth,trim={42.3cm 15.4cm 11cm 23.9cm}, clip]{figures/case_study.pdf} & \includegraphics[width=0.15\linewidth,trim={53.3cm 15.4cm 0.8cm 23.9cm}, clip]{figures/case_study.pdf}\\
        & Reference & Sim: 0.74 & Sim: 0.63 &  Sim: 0.47 & Sim: 0.46\\
        \rotatebox{90}{3M-Diffusion}&  \includegraphics[width=0.11\linewidth,trim={0cm 1cm 0cm 1.1cm}, clip]{figures/reference.png}  &  \includegraphics[width=0.15\linewidth,trim={19.3cm 5cm 34.8cm 35cm}, clip]{figures/case_study.pdf} & \includegraphics[width=0.15\linewidth,trim={29.3cm 4.7cm 23cm 35.9cm}, clip]{figures/case_study.pdf} & \includegraphics[width=0.15\linewidth,trim={42cm 4.3cm 12cm 34.9cm}, clip]{figures/case_study.pdf} & \includegraphics[width=0.15\linewidth,trim={53cm 4cm 0.8cm 34.9cm}, clip]{figures/case_study.pdf}\\
        & Reference & Sim: 0.87 & Sim: 0.87 & Sim: 0.87 & Sim: 0.86\\
        
    \end{tabular}
    }
    \caption{Comparative analysis of generated molecules from Mol-CADiff, 3M-Diffusion, and MolT5-large models under specified conditions (best viewed when zoomed in). The top 5 molecules are selected based on desired properties, with insolubility measured by logP (higher values indicate better insolubility). Reference molecules serve as ground truth.}
    \label{tab:case}
\end{table*}
\paragraph{Experimental Settings:}

For the graph encoder, we employ GIN~\cite{hu2019strategies} to obtain latent representations of molecular graphs, while Sci-BERT~\citep{beltagy2019scibert} is used for compressed text representation and HierVAE~\citep{jin2020hierarchical} is used as graph decoder. In the CaDiff module, we integrate graph and text information autoregressively, ensuring proper causal dependencies. For both conditional and unconditional generation tasks, we set the diffusion timesteps to $T=100$ during training, with $S_T=50$ for sampling in conditional generation and $25$ for unconditional generation on the ChEBI-20 dataset. The model is optimized using the Adam optimizer~\citep{jimmy2014adam} with an initial learning rate of $0.001$, training for 1000 epochs.

To incorporate classifier-free guidance~\citep{ho2022classifier}, we randomly drop the conditional embeddings with a probability of $0.1$ during training and inference for conditional generation, while for unconditional generation, this probability is set to $1.0$, effectively removing all contextual information. Additional settings, including the number of attention blocks and the proportion of clean tokens, are analyzed in the ablation study. Our implementation is based on PyTorch~\citep{paszke2019pytorch}, and all experiments were conducted on NVIDIA A100 GPUs.

\subsection{Quantitative Comparison}
For quantitative evaluation, we report the results of conditional generation in Table~\ref{tab:cond_all}. Our method consistently outperforms all SOTA baselines in overall performance by a large margin. In addition to excelling in overall metrics, Mol-CaDiff maintains strong similarity across all four datasets. Notably, it achieves substantial improvements in Novelty and Diversity, demonstrating its effectiveness in generating chemically valid, novel, and diverse molecules while preserving similarity to the given condition.

For unconditional generation, following 3M-Diffusion, we present results on the ChEBI-20 and PubChem datasets in Table~\ref{tab:uncond}. Here too, Mol-CaDiff achieves the highest overall performance, surpassing baseline methods. Furthermore, it exhibits strong Uniqueness, Novelty, KL Divergence, and FCD, either outperforming baselines or maintaining competitive results.
These results reinforce Mol-CADiff's superiority in both conditional and unconditional molecular generation, demonstrating its robustness, efficiency, and ability to generate effective molecular structures.

\subsection{Qualitative Comparison}

We present qualitative results demonstrating the efficacy of our method in Figure~\ref{fig:qual}, where we visualize five randomly sampled molecules generated by Mol-CADiff alongside their corresponding ground truth structures. It is evident that the generated molecules closely resemble the ground truth, faithfully capturing the textual descriptions provided. This visual similarity underscores our method’s ability to effectively leverage textual conditions, resulting in chemically valid molecules that reflect the desired properties.

Furthermore, we conduct detailed case studies in Table~\ref{tab:case}, comparing Mol-CADiff with two best-performing state-of-the-art methods, namely 3M-Diffusion, and MolT5-large, under specific textual instructions. As illustrated, Mol-CADiff consistently outperforms both baselines by generating molecules that better satisfy the given custom conditions. 

For instance, when instructed to generate molecules insoluble in water, Mol-CADiff produces molecules with higher logP values, clearly surpassing the baselines. Similarly, when generating molecules belonging to the class of flavones, our method achieves higher structural similarity to the reference molecules compared to other methods. These qualitative analyses further validate Mol-CADiff's superiority in generating high-quality, conditionally accurate molecules. 

Additional results, including more generated molecules and extended case studies, are provided in supplementary.

\section{Ablation Study}
In this section, we analyze the impact of key components in our method through ablation studies. To isolate the effect of each parameter or component, we vary one factor at a time while keeping all others fixed. We conduct our ablation study mostly on the PubChem dataset, as it offers a balanced trade-off between dataset size and computational efficiency.
\subsection{AR Step Decay}
\label{sub:decay}
\begin{figure}[t]
    \centering
    \includegraphics[width=0.65\linewidth]{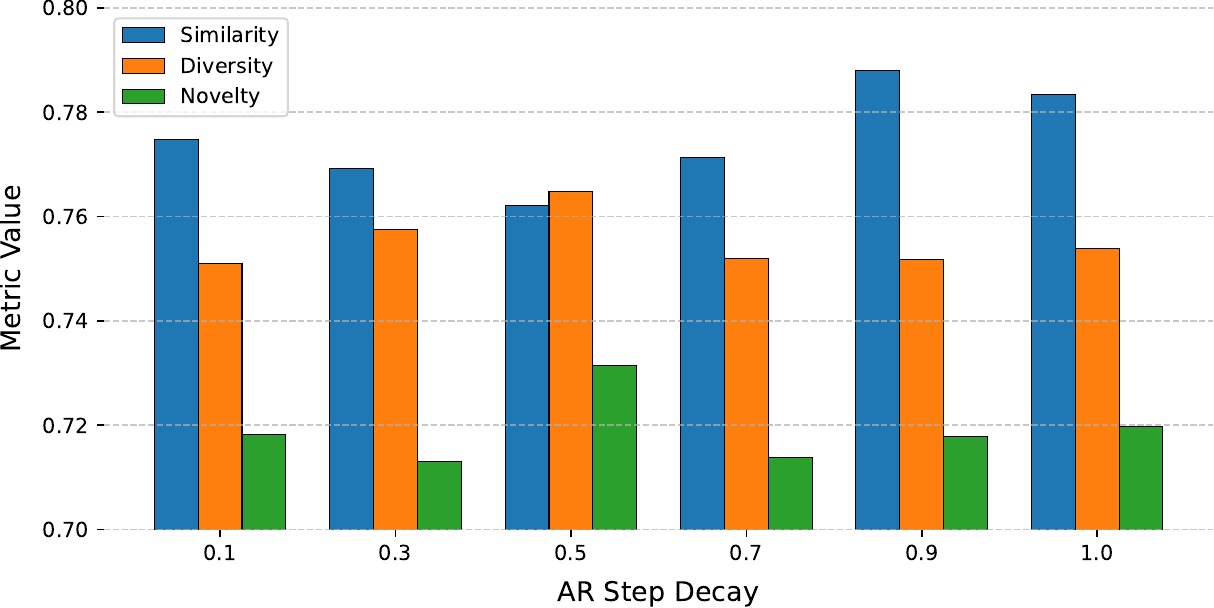}
    \caption{Ablation on AR Step Decay on PubChem dataset.}
    \label{fig:decay_abl}
\end{figure}

As discussed in Subsection~\ref{sub:method_ar_step}, the AR step sizes depend on the decay factor $\gamma$. Figure~\ref{fig:decay_abl} shows its impact on Similarity, Diversity, and Novelty in the PubChem dataset. Similarity remains relatively stable but improves slightly for higher values of $\gamma$, suggesting that larger AR steps help maintain structural coherence. Diversity peaks around $\gamma = 0.5$ but declines at extreme values, indicating that a balanced step size prevents excessive fragmentation or oversimplification. Novelty also peaks at $\gamma = 0.5$, showing that moderate decay encourages the exploration of new molecular structures.
\subsection{Effect of Partial Graph Latent Representation}

\label{sub:abl_full_part}
\begin{figure}[t]
    \centering
    \begin{subfigure}[t]{0.45\linewidth}
        \centering        
        \includegraphics[width=\linewidth]{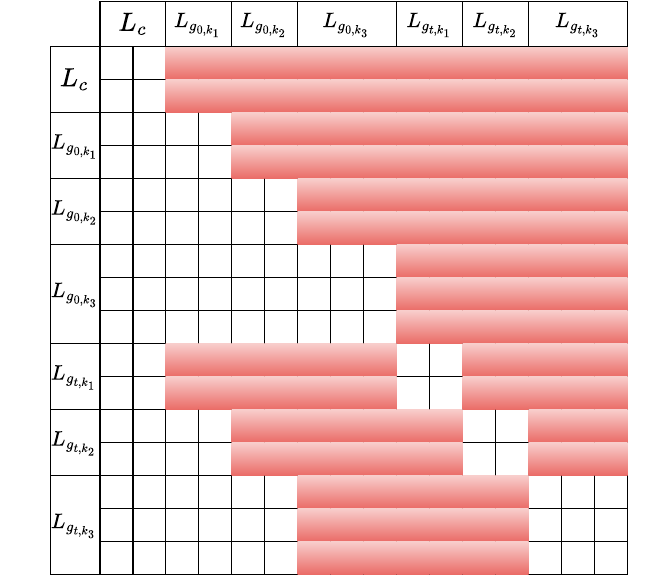}
        \caption{Full clean latent $L_{g_0}$}
    \end{subfigure}
    \hfill
    \begin{subfigure}[t]{0.45\linewidth}
        \centering
        \includegraphics[width=\linewidth]{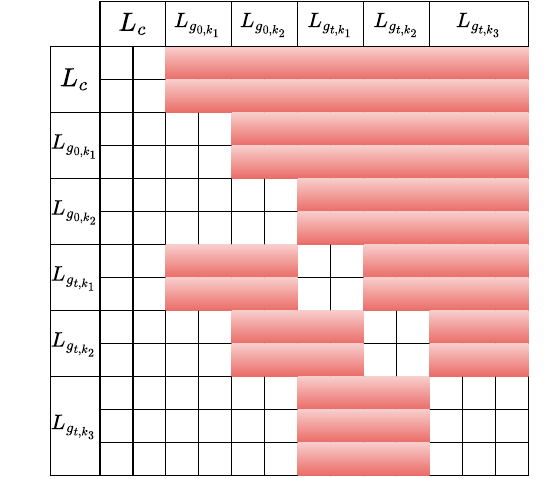}
        \caption{Partial clean latent $L_{g_0}$}
    \end{subfigure}
    \caption{Attention mask formation for full vs partial $L_{g_0}$}
    \label{fig:full_part}
\end{figure}

We analyze the impact of including the full vs. partial graph latent representation $L_{g_0}$. Figure~\ref{fig:full_part} illustrates an example attention mask for three AR steps, with token sizes of 2, 2, and 3. The left side represents the full latent $L_{g_0}$, which incorporates $L_{g_{0,k_3}}$, while the right side depicts the partial setting, where only $L_{g_{0,k_1}}$ and $L_{g_{0,k_2}}$ are included. Table~\ref{tab:abl_full_part} presents the results across all datasets. On ChEBI-20 and PubChem , the partial setting yields better overall performance compared to full inclusion. However, for MoMu, similarity slightly improves under the full setting, while for PCDes, full inclusion achieves marginal gains in novelty and diversity. Despite these variations, partial inclusion of $L_g$ offers computational efficiency by reducing token processing. Considering the trade-offs between performance and efficiency, we adopt the partial $L_{g_0}$ setting.
\begin{table}[t]
\centering
\caption{Full and Partial settings across all datasets.}
\label{tab:abl_full_part}
\resizebox{\linewidth}{!}{
\begin{threeparttable}
\begin{tabular}{l | c c c c c | c c c c c}
    \toprule
    & \multicolumn{5}{c|}{\shortstack[c]{\textbf{ChEBI-20}}} & \multicolumn{5}{c}{\shortstack[c]{\textbf{PubChem}}} \\
    Setting & Sim.& Nov.& Div.& Val.& Ove.& Sim. & Nov. & Div. & Val. & Ove.\\
    \midrule
    \textbf{Partial} & 81.92 & 67.35 & 75.69 & 100.0 & \textbf{81.24} & 78.35 & 71.97 & 75.39 & 100.0 & \textbf{81.43} \\
    \textbf{Full} & 82.56 & 64.81 & 71.75 & 100.0 & 79.78 & 76.35 & 73.60 & 75.65 & 100.0 & 81.40 \\
    \midrule
    & \multicolumn{5}{c|}{\shortstack[c]{\textbf{PCDes}}} & \multicolumn{5}{c}{\shortstack[c]{\textbf{MoMu}}} \\
    Setting & Sim.& Nov.& Div.& Val.& Ove.& Sim. & Nov. & Div. & Val. & Ove.\\
    \midrule
    \textbf{Partial} & 71.41 & 74.77 & 75.45 & 100.0 & 80.41 & 24.41 & 98.34 & 83.30 & 100.0 & 76.51 \\
    \textbf{Full} & 70.72 & 76.28 &76.07 & 100.0 & \textbf{80.77} & 25.56 & 98.02 & 82.98 & 100.0 & \textbf{76.64} \\
    \bottomrule
\end{tabular}
\end{threeparttable}
}
\end{table}

\subsection{Attention Blocks}
\begin{figure}[t]
    \centering
    \includegraphics[width=0.65\linewidth]{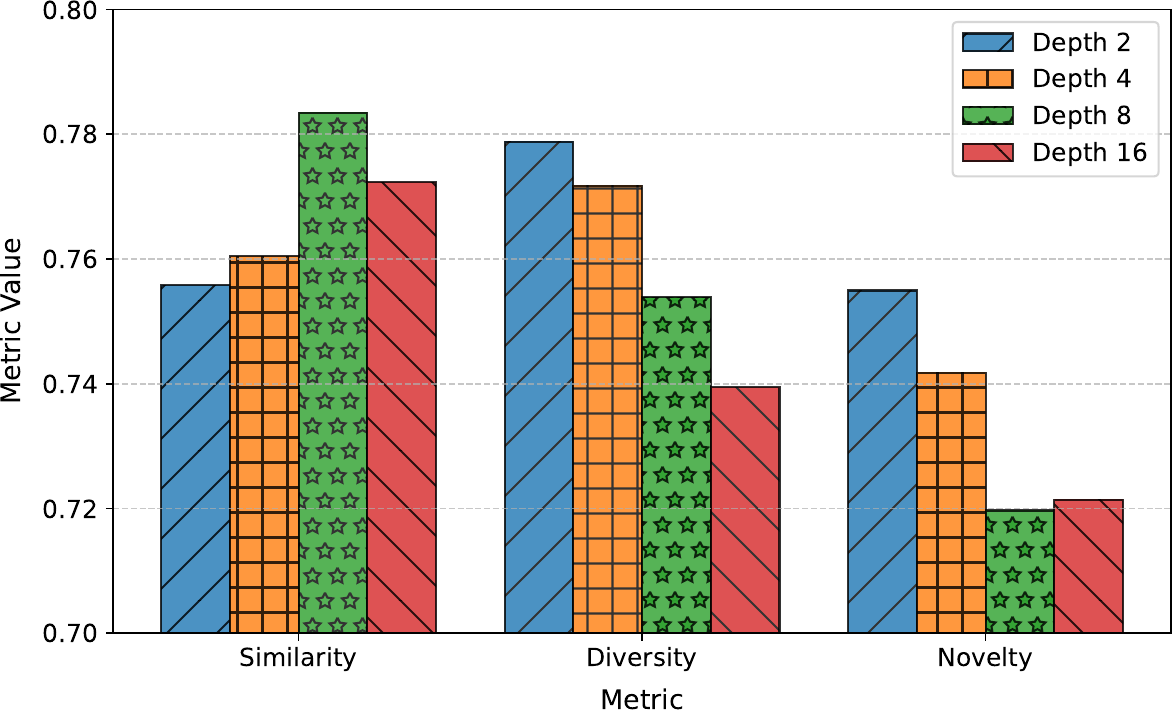}
    \caption{Effect of number of attention blocks}
    \label{fig:blocks}
\end{figure}

We analyze the impact of attention blocks (2, 4, 8, 16) on Similarity, Diversity, and Novelty in the PubChem dataset, excluding Validity (always 100.0\%). See Figure~\ref{fig:blocks}.

Similarity improves with more attention blocks, peaking at 8, suggesting that deeper architectures enhance molecular coherence. However, Diversity and Novelty decline beyond 2 blocks, indicating that excessive depth may reduce structural variation. To balance performance and efficiency, we set 8 blocks as the default, as increasing depth further raises the parameter count and computational cost.
\subsection{Sampling Time Steps}
\begin{figure}[t]
    \centering
    \includegraphics[width=0.65\linewidth]{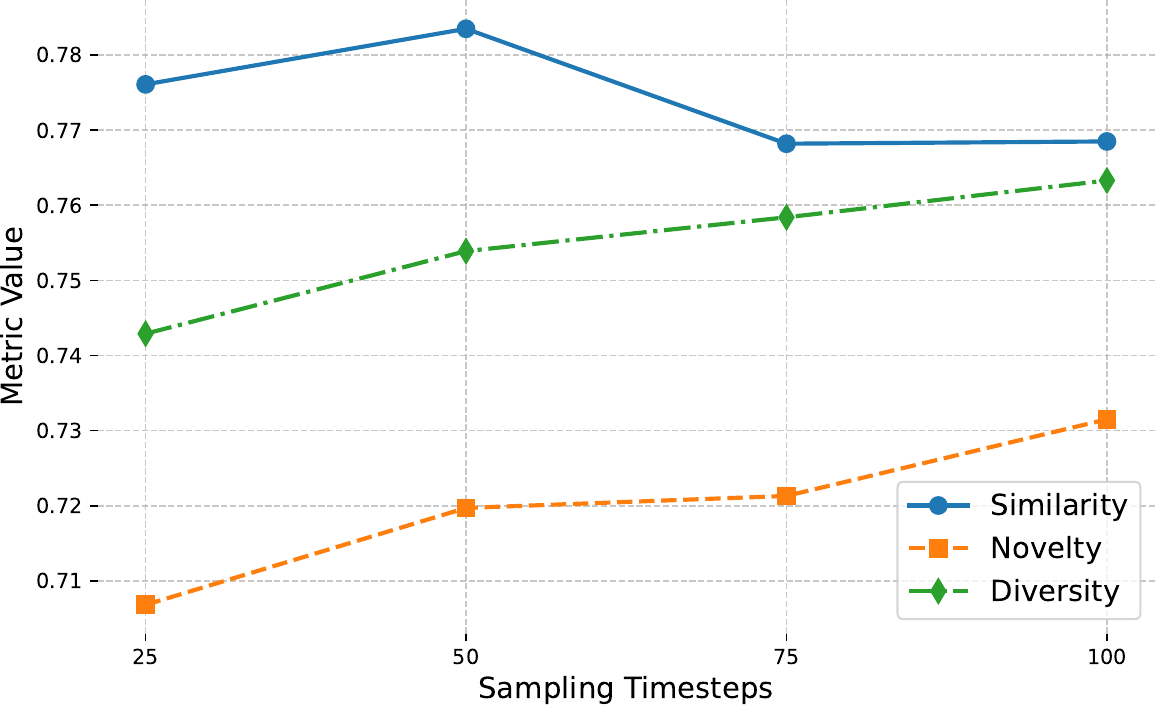}
    \caption{Ablation on sampling timesteps $S_T$.}
    \label{fig:sample_steps}
\end{figure}

We analyze the impact of different sampling timesteps ($S_T$) during inference on the quality of generated molecules. Specifically, we evaluate Similarity, Novelty, and Diversity across:
$ S_T \in \{25, 50, 75, 100\} $. As illustrated in Figure~\ref{fig:sample_steps}, Similarity is maximized at $S_T = 50$, suggesting that an adequate number of sampling steps improves the structural accuracy of generated molecules. Novelty and Diversity exhibit an increasing trend as $S_T$ increases, with the highest values observed at $S_T = 100$. This indicates that a greater number of sampling steps introduces more novel and diverse molecules. However, beyond $S_T = 50$, Similarity slightly decreases, likely due to excessive noise reduction leading to potential loss of structural integrity. Based on these observations, we set: $ S_T = 50 $ as the default configuration for conditional generation, as it balances Similarity, Novelty, and Diversity while maintaining computational efficiency.
\section{Conclusion}
We introduced \textbf{Mol-CADiff}, a framework that integrates \textit{autoregression with diffusion} through \textit{causal attention} for conditional and unconditional molecular generation. Unlike prior methods, our approach enforces structured dependencies, enabling more controlled and precise molecule generation. Experimental results demonstrate improvements in novelty, diversity, and validity while maintaining similarity. While Mol-CADiff enhances generation quality, future work can focus on improving efficiency and scalability for large-scale applications.

\section{Acknowledgment}
This research is supported in part by the NSF under Grant IIS 2327113 and ITE 2433190 and the NIH under Grants R21AG070909 and P30AG072946. We would like to thank NSF for support for the AI research resource with NAIRR240219. We thank the University of Kentucky Center for Computational Sciences and Information Technology Services Research Computing for their support and use of the Lipscomb Compute Cluster and associated research computing resources. We also acknowledge and thank those who created, cleaned, and curated the datasets used in this study.
\small
\bibliographystyle{ieeenat_fullname}
\bibliography{bibliography}

\begin{thebibliography}{51}
\providecommand{\natexlab}[1]{#1}
\providecommand{\url}[1]{\texttt{#1}}
\expandafter\ifx\csname urlstyle\endcsname\relax
  \providecommand{\doi}[1]{doi: #1}\else
  \providecommand{\doi}{doi: \begingroup \urlstyle{rm}\Url}\fi

\bibitem[Beltagy et~al.(2019)Beltagy, Lo, and Cohan]{beltagy2019scibert}
Iz Beltagy, Kyle Lo, and Arman Cohan.
\newblock Scibert: A pretrained language model for scientific text.
\newblock \emph{arXiv preprint arXiv:1903.10676}, 2019.

\bibitem[Bjerrum and Threlfall(2017)]{bjerrum2017molecular}
Esben~Jannik Bjerrum and Richard Threlfall.
\newblock Molecular generation with recurrent neural networks (rnns).
\newblock \emph{arXiv preprint arXiv:1705.04612}, 2017.

\bibitem[Blum and Reymond(2009)]{blum2009970}
Lorenz~C Blum and Jean-Louis Reymond.
\newblock 970 million druglike small molecules for virtual screening in the chemical universe database gdb-13.
\newblock \emph{Journal of the American Chemical Society}, 131\penalty0 (25):\penalty0 8732--8733, 2009.

\bibitem[Brown et~al.(2019)Brown, Fiscato, Segler, and Vaucher]{brown2019guacamol}
Nathan Brown, Marco Fiscato, Marwin~HS Segler, and Alain~C Vaucher.
\newblock Guacamol: benchmarking models for de novo molecular design.
\newblock \emph{Journal of chemical information and modeling}, 59\penalty0 (3):\penalty0 1096--1108, 2019.

\bibitem[Christofidellis et~al.(2023)Christofidellis, Giannone, Born, Winther, Laino, and Manica]{christofidellis2023unifying}
Dimitrios Christofidellis, Giorgio Giannone, Jannis Born, Ole Winther, Teodoro Laino, and Matteo Manica.
\newblock Unifying molecular and textual representations via multi-task language modelling.
\newblock \emph{arXiv preprint arXiv:2301.12586}, 2023.

\bibitem[Deng et~al.(2024)Deng, Zh, Li, Guan, and Fan]{deng2024causal}
Chaorui Deng, Deyao Zh, Kunchang Li, Shi Guan, and Haoqi Fan.
\newblock Causal diffusion transformers for generative modeling.
\newblock \emph{arXiv preprint arXiv:2412.12095}, 2024.

\bibitem[Diamant et~al.(2023)Diamant, Tseng, Chuang, Biancalani, and Scalia]{diamant2023improving}
Nathaniel~Lee Diamant, Alex~M Tseng, Kangway~V Chuang, Tommaso Biancalani, and Gabriele Scalia.
\newblock Improving graph generation by restricting graph bandwidth.
\newblock In \emph{International Conference on Machine Learning}, pages 7939--7959. PMLR, 2023.

\bibitem[Durant et~al.(2002)Durant, Leland, Henry, and Nourse]{durant2002reoptimization}
Joseph~L Durant, Burton~A Leland, Douglas~R Henry, and James~G Nourse.
\newblock Reoptimization of mdl keys for use in drug discovery.
\newblock \emph{Journal of chemical information and computer sciences}, 42\penalty0 (6):\penalty0 1273--1280, 2002.

\bibitem[Edwards et~al.(2021)Edwards, Zhai, and Ji]{edwards2021text2mol}
Carl Edwards, ChengXiang Zhai, and Heng Ji.
\newblock Text2mol: Cross-modal molecule retrieval with natural language queries.
\newblock In \emph{Proceedings of the 2021 Conference on Empirical Methods in Natural Language Processing}, pages 595--607, 2021.

\bibitem[Edwards et~al.(2022)Edwards, Lai, Ros, Honke, Cho, and Ji]{edwards2022translation}
Carl Edwards, Tuan Lai, Kevin Ros, Garrett Honke, Kyunghyun Cho, and Heng Ji.
\newblock Translation between molecules and natural language.
\newblock \emph{arXiv preprint arXiv:2204.11817}, 2022.

\bibitem[Fang et~al.(2023{\natexlab{a}})Fang, Liang, Zhang, Liu, Huang, Chen, Fan, and Chen]{fang2023mol}
Yin Fang, Xiaozhuan Liang, Ningyu Zhang, Kangwei Liu, Rui Huang, Zhuo Chen, Xiaohui Fan, and Huajun Chen.
\newblock Mol-instructions: A large-scale biomolecular instruction dataset for large language models.
\newblock \emph{arXiv preprint arXiv:2306.08018}, 2023{\natexlab{a}}.

\bibitem[Fang et~al.(2023{\natexlab{b}})Fang, Zhang, Chen, Fan, and Chen]{fang2023molecular}
Yin Fang, Ningyu Zhang, Zhuo Chen, Xiaohui Fan, and Huajun Chen.
\newblock Molecular language model as multi-task generator.
\newblock \emph{arXiv preprint arXiv:2301.11259}, 2023{\natexlab{b}}.

\bibitem[Flam-Shepherd et~al.(2022)Flam-Shepherd, Zhu, and Aspuru-Guzik]{flam2022language}
Daniel Flam-Shepherd, Kevin Zhu, and Al{\'a}n Aspuru-Guzik.
\newblock Language models can learn complex molecular distributions.
\newblock \emph{Nature Communications}, 13\penalty0 (1):\penalty0 3293, 2022.

\bibitem[G{\'o}mez-Bombarelli et~al.(2018)G{\'o}mez-Bombarelli, Wei, Duvenaud, Hern{\'a}ndez-Lobato, S{\'a}nchez-Lengeling, Sheberla, Aguilera-Iparraguirre, Hirzel, Adams, and Aspuru-Guzik]{gomez2018automatic}
Rafael G{\'o}mez-Bombarelli, Jennifer~N Wei, David Duvenaud, Jos{\'e}~Miguel Hern{\'a}ndez-Lobato, Benjam{\'\i}n S{\'a}nchez-Lengeling, Dennis Sheberla, Jorge Aguilera-Iparraguirre, Timothy~D Hirzel, Ryan~P Adams, and Al{\'a}n Aspuru-Guzik.
\newblock Automatic chemical design using a data-driven continuous representation of molecules.
\newblock \emph{ACS central science}, 4\penalty0 (2):\penalty0 268--276, 2018.

\bibitem[Goyal et~al.(2020)Goyal, Jain, and Ranu]{goyal2020graphgen}
Nikhil Goyal, Harsh~Vardhan Jain, and Sayan Ranu.
\newblock Graphgen: A scalable approach to domain-agnostic labeled graph generation.
\newblock In \emph{Proceedings of The Web Conference 2020}, pages 1253--1263, 2020.

\bibitem[Hajduk and Greer(2007)]{hajduk2007decade}
Philip~J Hajduk and Jonathan Greer.
\newblock A decade of fragment-based drug design: strategic advances and lessons learned.
\newblock \emph{Nature reviews Drug discovery}, 6\penalty0 (3):\penalty0 211--219, 2007.

\bibitem[Ho and Salimans(2022)]{ho2022classifier}
Jonathan Ho and Tim Salimans.
\newblock Classifier-free diffusion guidance.
\newblock \emph{arXiv preprint arXiv:2207.12598}, 2022.

\bibitem[Ho et~al.(2020)Ho, Jain, and Abbeel]{ho2020denoising}
Jonathan Ho, Ajay Jain, and Pieter Abbeel.
\newblock Denoising diffusion probabilistic models.
\newblock \emph{Advances in neural information processing systems}, 33:\penalty0 6840--6851, 2020.

\bibitem[Hu et~al.(2019)Hu, Liu, Gomes, Zitnik, Liang, Pande, and Leskovec]{hu2019strategies}
Weihua Hu, Bowen Liu, Joseph Gomes, Marinka Zitnik, Percy Liang, Vijay Pande, and Jure Leskovec.
\newblock Strategies for pre-training graph neural networks.
\newblock \emph{arXiv preprint arXiv:1905.12265}, 2019.

\bibitem[Irwin et~al.(2012)Irwin, Sterling, Mysinger, Bolstad, and Coleman]{irwin2012zinc}
John~J Irwin, Teague Sterling, Michael~M Mysinger, Erin~S Bolstad, and Ryan~G Coleman.
\newblock Zinc: a free tool to discover chemistry for biology.
\newblock \emph{Journal of chemical information and modeling}, 52\penalty0 (7):\penalty0 1757--1768, 2012.

\bibitem[Jimmy and Diederik(2014)]{jimmy2014adam}
Ba Jimmy and P Diederik.
\newblock Adam: A method for stochastic optimization.
\newblock \emph{arXiv preprint arXiv: 1412.6980}, page 2014, 2014.

\bibitem[Jin et~al.(2018)Jin, Barzilay, and Jaakkola]{jin2018junction}
Wengong Jin, Regina Barzilay, and Tommi Jaakkola.
\newblock Junction tree variational autoencoder for molecular graph generation.
\newblock In \emph{International conference on machine learning}, pages 2323--2332. PMLR, 2018.

\bibitem[Jin et~al.(2020)Jin, Barzilay, and Jaakkola]{jin2020hierarchical}
Wengong Jin, Regina Barzilay, and Tommi Jaakkola.
\newblock Hierarchical generation of molecular graphs using structural motifs.
\newblock In \emph{International conference on machine learning}, pages 4839--4848. PMLR, 2020.

\bibitem[Jo et~al.(2022)Jo, Lee, and Hwang]{jo2022score}
Jaehyeong Jo, Seul Lee, and Sung~Ju Hwang.
\newblock Score-based generative modeling of graphs via the system of stochastic differential equations.
\newblock In \emph{International Conference on Machine Learning}, pages 10362--10383. PMLR, 2022.

\bibitem[Kingma and Welling(2013)]{kingma2013auto}
Diederik~P Kingma and Max Welling.
\newblock Auto-encoding variational bayes.
\newblock \emph{arXiv preprint arXiv:1312.6114}, 2013.

\bibitem[Kong et~al.(2022)Kong, Huang, Tan, and Liu]{kong2022molecule}
Xiangzhe Kong, Wenbing Huang, Zhixing Tan, and Yang Liu.
\newblock Molecule generation by principal subgraph mining and assembling.
\newblock \emph{Advances in Neural Information Processing Systems}, 35:\penalty0 2550--2563, 2022.

\bibitem[Krenn et~al.(2022)Krenn, Ai, Barthel, Carson, Frei, Frey, Friederich, Gaudin, Gayle, Jablonka, et~al.]{krenn2022selfies}
Mario Krenn, Qianxiang Ai, Senja Barthel, Nessa Carson, Angelo Frei, Nathan~C Frey, Pascal Friederich, Th{\'e}ophile Gaudin, Alberto~Alexander Gayle, Kevin~Maik Jablonka, et~al.
\newblock Selfies and the future of molecular string representations.
\newblock \emph{Patterns}, 3\penalty0 (10), 2022.

\bibitem[Li et~al.(2018)Li, Vinyals, Dyer, Pascanu, and Battaglia]{li2018learning}
Yujia Li, Oriol Vinyals, Chris Dyer, Razvan Pascanu, and Peter Battaglia.
\newblock Learning deep generative models of graphs.
\newblock \emph{arXiv preprint arXiv:1803.03324}, 2018.

\bibitem[Liao et~al.(2019)Liao, Li, Song, Wang, Hamilton, Duvenaud, Urtasun, and Zemel]{liao2019efficient}
Renjie Liao, Yujia Li, Yang Song, Shenlong Wang, Will Hamilton, David~K Duvenaud, Raquel Urtasun, and Richard Zemel.
\newblock Efficient graph generation with graph recurrent attention networks.
\newblock \emph{Advances in neural information processing systems}, 32, 2019.

\bibitem[Liu et~al.(2018)Liu, Allamanis, Brockschmidt, and Gaunt]{liu2018constrained}
Qi Liu, Miltiadis Allamanis, Marc Brockschmidt, and Alexander Gaunt.
\newblock Constrained graph variational autoencoders for molecule design.
\newblock \emph{Advances in neural information processing systems}, 31, 2018.

\bibitem[Liu et~al.(2023)Liu, Li, Luo, Fei, Cao, Kawaguchi, Wang, and Chua]{liu2023molca}
Zhiyuan Liu, Sihang Li, Yanchen Luo, Hao Fei, Yixin Cao, Kenji Kawaguchi, Xiang Wang, and Tat-Seng Chua.
\newblock Molca: Molecular graph-language modeling with cross-modal projector and uni-modal adapter.
\newblock \emph{arXiv preprint arXiv:2310.12798}, 2023.

\bibitem[Luo et~al.(2021)Luo, Yan, and Ji]{luo2021graphdf}
Youzhi Luo, Keqiang Yan, and Shuiwang Ji.
\newblock Graphdf: A discrete flow model for molecular graph generation.
\newblock In \emph{International Conference on Machine Learning}, pages 7192--7203. PMLR, 2021.

\bibitem[Madhawa et~al.(2019)Madhawa, Ishiguro, Nakago, and Abe]{madhawa2019graphnvp}
Kaushalya Madhawa, Katushiko Ishiguro, Kosuke Nakago, and Motoki Abe.
\newblock Graphnvp: An invertible flow model for generating molecular graphs.
\newblock \emph{arXiv preprint arXiv:1905.11600}, 2019.

\bibitem[Makhzani et~al.(2015)Makhzani, Shlens, Jaitly, Goodfellow, and Frey]{makhzani2015adversarial}
Alireza Makhzani, Jonathon Shlens, Navdeep Jaitly, Ian Goodfellow, and Brendan Frey.
\newblock Adversarial autoencoders.
\newblock \emph{arXiv preprint arXiv:1511.05644}, 2015.

\bibitem[Paszke et~al.(2019)Paszke, Gross, Massa, Lerer, Bradbury, Chanan, Killeen, Lin, Gimelshein, Antiga, et~al.]{paszke2019pytorch}
Adam Paszke, Sam Gross, Francisco Massa, Adam Lerer, James Bradbury, Gregory Chanan, Trevor Killeen, Zeming Lin, Natalia Gimelshein, Luca Antiga, et~al.
\newblock Pytorch: An imperative style, high-performance deep learning library.
\newblock \emph{Advances in neural information processing systems}, 32, 2019.

\bibitem[Popova et~al.(2019)Popova, Shvets, Oliva, and Isayev]{popova2019molecularrnn}
Mariya Popova, Mykhailo Shvets, Junier Oliva, and Olexandr Isayev.
\newblock Molecularrnn: Generating realistic molecular graphs with optimized properties.
\newblock \emph{arXiv preprint arXiv:1905.13372}, 2019.

\bibitem[Preuer et~al.(2018)Preuer, Renz, Unterthiner, Hochreiter, and Klambauer]{preuer2018frechet}
Kristina Preuer, Philipp Renz, Thomas Unterthiner, Sepp Hochreiter, and Gunter Klambauer.
\newblock Fr{\'e}chet chemnet distance: a metric for generative models for molecules in drug discovery.
\newblock \emph{Journal of chemical information and modeling}, 58\penalty0 (9):\penalty0 1736--1741, 2018.

\bibitem[Prykhodko et~al.(2019)Prykhodko, Johansson, Kotsias, Ar{\'u}s-Pous, Bjerrum, Engkvist, and Chen]{prykhodko2019novo}
Oleksii Prykhodko, Simon~Viet Johansson, Panagiotis-Christos Kotsias, Josep Ar{\'u}s-Pous, Esben~Jannik Bjerrum, Ola Engkvist, and Hongming Chen.
\newblock A de novo molecular generation method using latent vector based generative adversarial network.
\newblock \emph{Journal of Cheminformatics}, 11:\penalty0 1--13, 2019.

\bibitem[Raffel et~al.(2020)Raffel, Shazeer, Roberts, Lee, Narang, Matena, Zhou, Li, and Liu]{raffel2020exploring}
Colin Raffel, Noam Shazeer, Adam Roberts, Katherine Lee, Sharan Narang, Michael Matena, Yanqi Zhou, Wei Li, and Peter~J Liu.
\newblock Exploring the limits of transfer learning with a unified text-to-text transformer.
\newblock \emph{The Journal of Machine Learning Research}, 21\penalty0 (1):\penalty0 5485--5551, 2020.

\bibitem[Rombach et~al.(2022)Rombach, Blattmann, Lorenz, Esser, and Ommer]{rombach2022high}
Robin Rombach, Andreas Blattmann, Dominik Lorenz, Patrick Esser, and Bj{\"o}rn Ommer.
\newblock High-resolution image synthesis with latent diffusion models.
\newblock In \emph{Proceedings of the IEEE/CVF conference on computer vision and pattern recognition}, pages 10684--10695, 2022.

\bibitem[Rupp et~al.(2012)Rupp, Tkatchenko, M{\"u}ller, and Von~Lilienfeld]{rupp2012fast}
Matthias Rupp, Alexandre Tkatchenko, Klaus-Robert M{\"u}ller, and O~Anatole Von~Lilienfeld.
\newblock Fast and accurate modeling of molecular atomization energies with machine learning.
\newblock \emph{Physical review letters}, 108\penalty0 (5):\penalty0 058301, 2012.

\bibitem[Segler et~al.(2018)Segler, Kogej, Tyrchan, and Waller]{segler2018generating}
Marwin~HS Segler, Thierry Kogej, Christian Tyrchan, and Mark~P Waller.
\newblock Generating focused molecule libraries for drug discovery with recurrent neural networks.
\newblock \emph{ACS central science}, 4\penalty0 (1):\penalty0 120--131, 2018.

\bibitem[Song et~al.(2020)Song, Meng, and Ermon]{song2020denoising}
Jiaming Song, Chenlin Meng, and Stefano Ermon.
\newblock Denoising diffusion implicit models.
\newblock \emph{arXiv preprint arXiv:2010.02502}, 2020.

\bibitem[Su et~al.(2022)Su, Du, Yang, Zhou, Li, Rao, Sun, Lu, and Wen]{su2022molecular}
Bing Su, Dazhao Du, Zhao Yang, Yujie Zhou, Jiangmeng Li, Anyi Rao, Hao Sun, Zhiwu Lu, and Ji-Rong Wen.
\newblock A molecular multimodal foundation model associating molecule graphs with natural language.
\newblock \emph{arXiv preprint arXiv:2209.05481}, 2022.

\bibitem[Vignac et~al.(2022)Vignac, Krawczuk, Siraudin, Wang, Cevher, and Frossard]{vignac2022digress}
Clement Vignac, Igor Krawczuk, Antoine Siraudin, Bohan Wang, Volkan Cevher, and Pascal Frossard.
\newblock Digress: Discrete denoising diffusion for graph generation.
\newblock \emph{arXiv preprint}, 2022.

\bibitem[Weininger(1988)]{weininger1988smiles}
David Weininger.
\newblock Smiles, a chemical language and information system. 1. introduction to methodology and encoding rules.
\newblock \emph{Journal of chemical information and computer sciences}, 28\penalty0 (1):\penalty0 31--36, 1988.

\bibitem[Xiao et~al.(2023)Xiao, Cui, Zhu, and Honavar]{MolBind2024}
Teng Xiao, Chao Cui, Huaisheng Zhu, and Vasant Honavar.
\newblock Molbind: Multimodal alignment of language, molecules, and proteins.
\newblock \emph{arXiv preprint}, 2023.

\bibitem[You et~al.(2018)You, Liu, Ying, Pande, and Leskovec]{you2018graph}
Jiaxuan You, Bowen Liu, Zhitao Ying, Vijay Pande, and Jure Leskovec.
\newblock Graph convolutional policy network for goal-directed molecular graph generation.
\newblock \emph{Advances in neural information processing systems}, 31, 2018.

\bibitem[Zang and Wang(2020)]{zang2020moflow}
Chengxi Zang and Fei Wang.
\newblock Moflow: an invertible flow model for generating molecuclar graphs.
\newblock In \emph{Proceedings of the 26th ACM SIGKDD international conference on knowledge discovery \& data mining}, pages 617--626, 2020.

\bibitem[Zeng et~al.(2022)Zeng, Yao, Liu, and Sun]{zeng2022deep}
Zheni Zeng, Yuan Yao, Zhiyuan Liu, and Maosong Sun.
\newblock A deep-learning system bridging molecule structure and biomedical text with comprehension comparable to human professionals.
\newblock \emph{Nature communications}, 13\penalty0 (1):\penalty0 862, 2022.

\bibitem[Zhu et~al.(2024)Zhu, Xiao, and Honavar]{zhu20243m}
Huaisheng Zhu, Teng Xiao, and Vasant~G Honavar.
\newblock 3m-diffusion: Latent multi-modal diffusion for language-guided molecular structure generation.
\newblock In \emph{First Conference on Language Modeling}, 2024.

\end{thebibliography}
\maketitlesupplementary
\setcounter{section}{0}
\setcounter{figure}{0}
\setcounter{table}{0}

\setcounter{algorithm}{0}

\renewcommand{\figurename}{S-Figure}
\renewcommand{\tablename}{S-Table}
\floatname{algorithm}{S-Algorithm}

This supplementary document provides additional details and experimental results that complement our main paper. We include an algorithm of autoregressive causal attention mask, extended qualitative results for conditionally and unconditionally generated molecules, and additional case studies comparing Mol-CADiff with other baselines. These materials further demonstrate the effectiveness of our proposed method. To distinguish materials in this supplementary from the main text, we add the prefix \textbf{S-} before Figures, Tables, and Algorithms.

\section{Autoregressive Causal Mask}
Our framework integrates a causal attention mechanism to model dependencies in an autoregressive manner, ensuring structured molecule generation. In this section, we provide the detailed algorithm for generating the causal attention mask, which controls how information flows during training. The algorithm enforces appropriate token dependencies, allowing partial clean graph latents and conditional text tokens to guide the denoising process effectively. The full procedure is detailed in S-Algorithm~\ref{alg:causal_mask}.
\begin{algorithm}[H]
\caption{Autoregressive Causal Attention Mask}
\label{alg:causal_mask}
\textbf{Input:} $l$ (sample length), $cl$ (conditional length), $as$ (split sizes), $cm$ (cumulative sum of split sizes)\\
\textbf{Output:} Causal attention mask $M$
\begin{algorithmic}[1]
\STATE $v \gets l - as[-1]$, $ctx \gets cl + v$, $seq \gets ctx + l$ 
\STATE Initialize $M \in \mathbb{R}^{seq \times seq}$ with ones
\STATE $M[:, :cl] \gets 0$ 

\STATE $T_1 \gets \mathbf{1}^{v \times v}$, $T_2 \gets \mathbf{1}^{l \times v}$, $T_3 \gets \mathbf{1}^{l \times l}$

\FOR{$i = 0$ to $|as| - 2$}
    \STATE $T_1[cm[i] : cm[i+1], 0 : cm[i+1]] \gets 0$
    \STATE $T_2[cm[i+1] : cm[i+2], 0 : cm[i+1]] \gets 0$
\ENDFOR
\FOR{$i = 0$ to $|as| - 1$}
    \STATE $T_3[cm[i] : cm[i+1], cm[i] : cm[i+1]] \gets 0$
\ENDFOR

\STATE $M[cl:ctx, cl:ctx] \gets T_1$, $M[ctx:, cl:ctx] \gets T_2$, $M[ctx:, ctx:] \gets T_3$

\STATE \textbf{return} $M$
\end{algorithmic}
\end{algorithm}

\section{Conditionally Generated Samples}
\begin{figure*}[h]
    \centering
    \includegraphics[width=\linewidth]{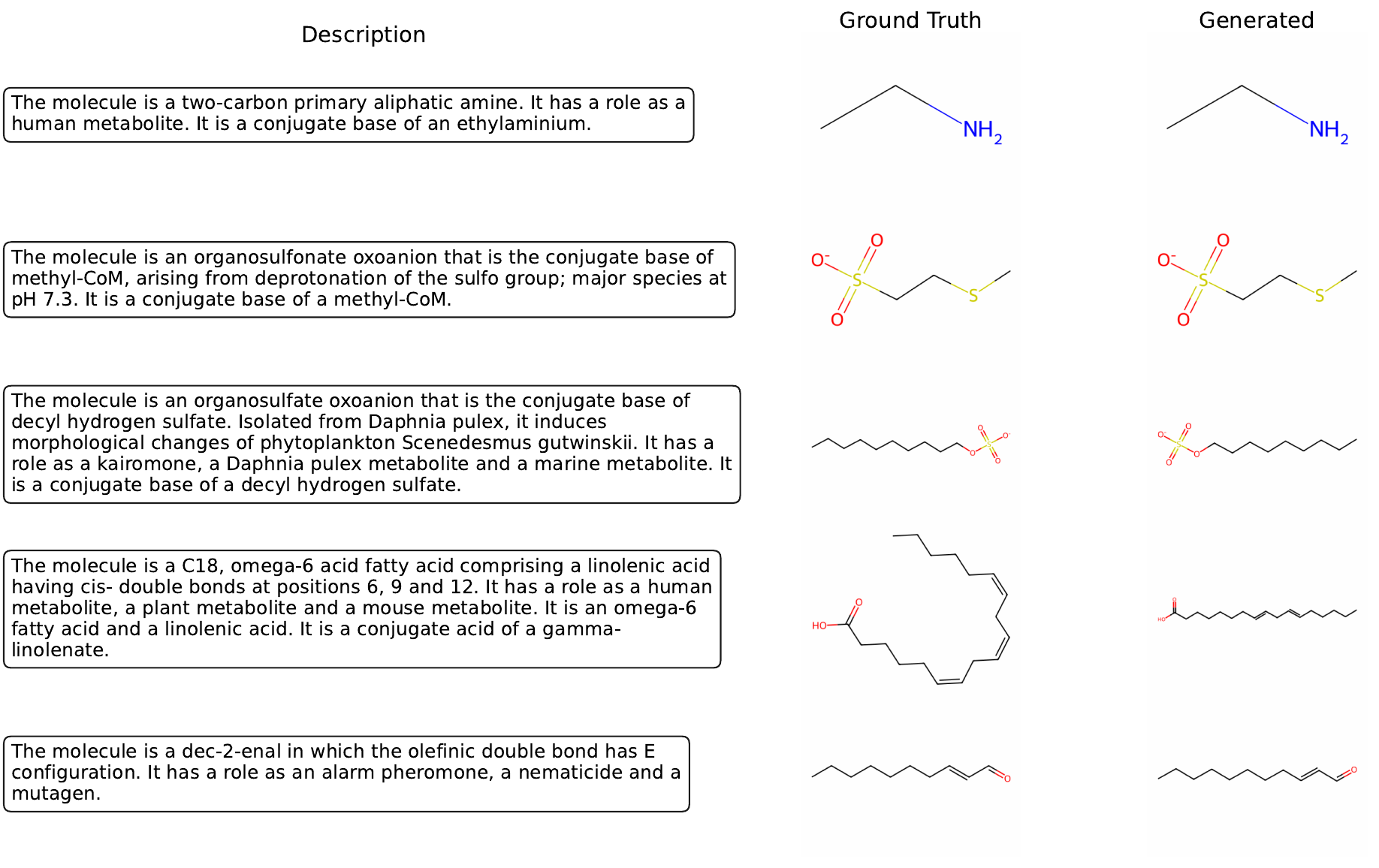}
    \caption{Generated molecules from ChEBI-20 test set, conditioned on text prompts. The results retain textual information and align closely with ground truth (best viewed when zoomed in).}
    \label{fig:more_conditional}
\end{figure*}

To further validate the effectiveness of Mol-CADiff in text-guided molecular generation, we present additional conditional samples. These molecules are generated based on text prompts from the ChEBI-20 test set, which were unseen during training. As shown in S-Figure~\ref{fig:more_conditional}, the generated molecules maintain high fidelity to the provided textual conditions while preserving structural validity. The results highlight the ability of our model to align molecular structures with textual descriptions in a chemically meaningful manner.

\section{Unconditionally Generated Samples}
\begin{figure*}[h]
    \centering
    \includegraphics[width=\linewidth]{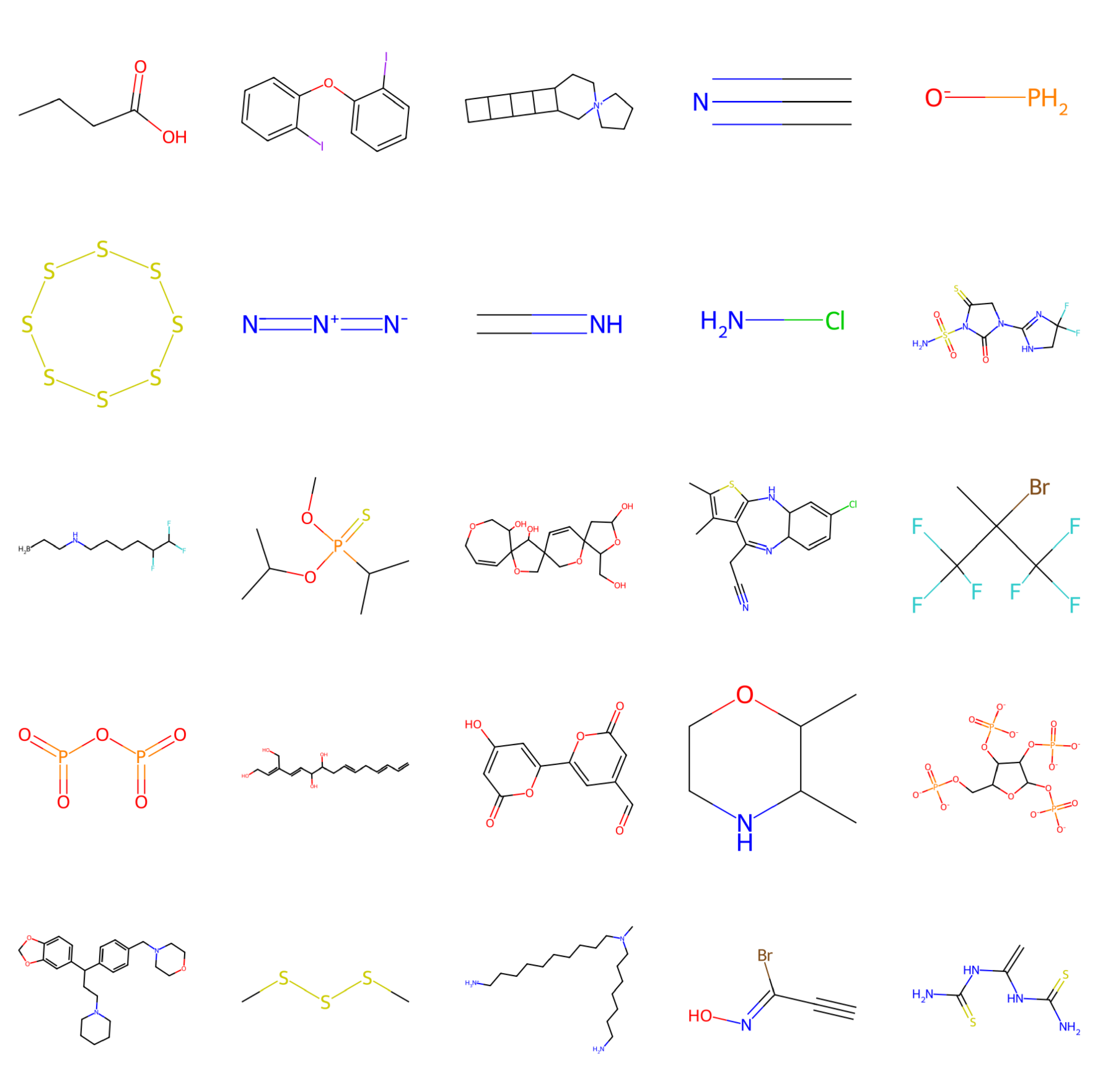}
    \caption{Unconditional molecular generation showcasing diverse and valid molecules.}
    \label{fig:unconditional}
\end{figure*}

In addition to conditional generation, Mol-CADiff is capable of generating molecules without textual prompts, relying purely on the learned distribution. S-Figure~\ref{fig:unconditional} presents diverse and valid molecules sampled unconditionally, demonstrating the flexibility of our framework in exploring the chemical space. These results confirm that Mol-CADiff is not only effective in controlled molecular generation but also robust in generating diverse and chemically plausible structures in an unconstrained setting.
\section{Additional Case Studies}
To further analyze the performance of Mol-CADiff, we compare its generated molecules with those from 3M-Diffusion and MolT5-large across different chemical property conditions. The tables in this section provide a qualitative comparison of the generated molecules under specific textual conditions. Each row contains molecules generated by different models, along with their respective QED (quantitative estimate of drug-likeness) or similarity scores.

\textbf{Drug-likeness Study}: S-Table~\ref{tab:case_drug} presents molecules generated under the condition \textit{"This molecule is like a drug."} Mol-CADiff achieves higher QED values compared to other models, indicating better alignment with drug-like properties.

\textbf{Non-Drug Molecule Study}: Conversely, in S-Table~\ref{tab:case_undrug}, molecules are generated under the opposite condition, \textit{"This molecule is not like a drug."} The lower QED values suggest that Mol-CADiff effectively learns to generate molecules outside the drug-like chemical space.

\textbf{Functional Group Matching}: S-Table~\ref{tab:case_an} showcases generated molecules under the condition \textit{"The molecule is an anthocyanidin cation."} The similarity scores indicate that Mol-CADiff achieves better structural alignment with the reference molecules compared to baselines.

These case studies further validate that Mol-CADiff not only generates valid and novel molecules but also effectively captures the nuanced properties required for text-conditional molecular generation.
\begin{table*}[!ht]
    \centering
    \resizebox{\linewidth}{!}{
    \begin{tabular}{lccccc}
        & \multicolumn{5}{c}{\textbf{Condition:} This molecule is like a drug.}\\
         \rotatebox{90}{Mol-CADiff}&  \includegraphics[width=0.15\linewidth,trim={0cm 0.8cm 0cm 1.1cm}, clip]{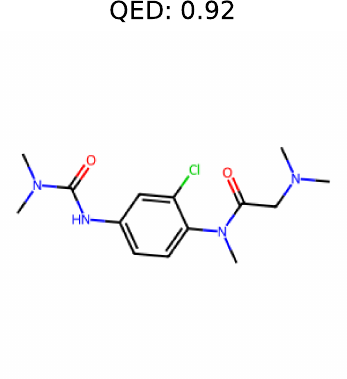}&  \includegraphics[width=0.15\linewidth,trim={0cm 0.8cm 0cm 1.1cm}, clip]{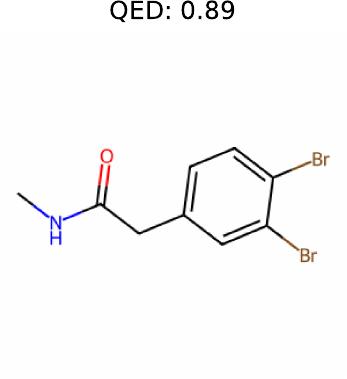}&  \includegraphics[width=0.15\linewidth,trim={0cm 1cm 0cm 1.1cm}, clip]{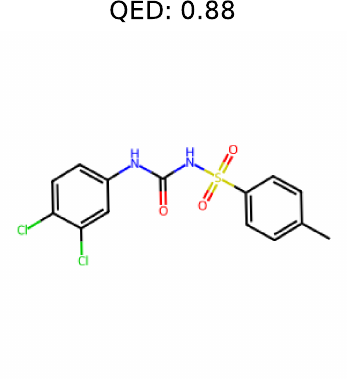}&  \includegraphics[width=0.15\linewidth,trim={0cm 0.8cm 0cm 1.1cm}, clip]{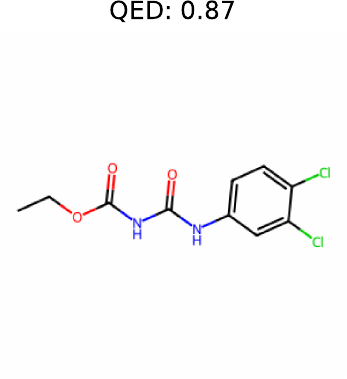}&  \includegraphics[width=0.15\linewidth,trim={0cm 0.8cm 0cm 1.1cm}, clip]{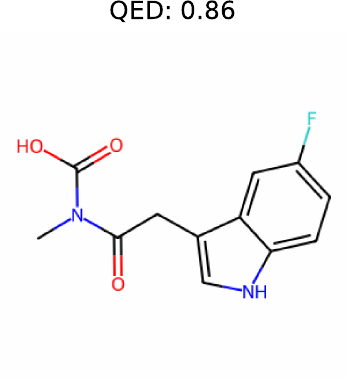}\\
         & QED: 0.92 & QED: 0.89 & QED: 0.88 & QED: 0.87 & QED: 0.86\\
        \rotatebox{90}{MolT5-large}&  \rotatebox{90}{\includegraphics[width=0.15\linewidth,trim={11cm 12cm 44cm 2.9cm}, clip]{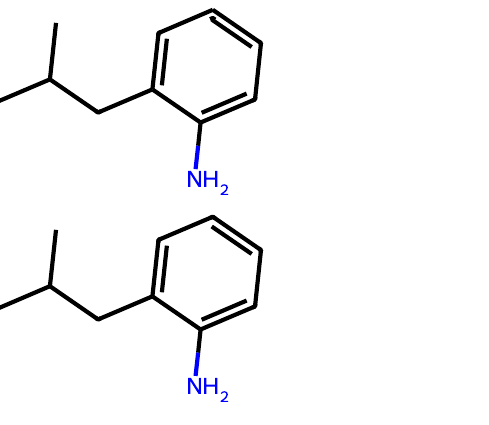}} & \rotatebox{90}{\includegraphics[width=0.15\linewidth,trim={22cm 12cm 34cm 2.9cm}, clip]{figures/case_drug.pdf}} & \includegraphics[width=0.15\linewidth,trim={30cm 13.5cm 22cm 2.9cm}, clip]{figures/case_drug.pdf} & \includegraphics[width=0.15\linewidth,trim={40cm 13.5cm 12cm 2.9cm}, clip]{figures/case_drug.pdf} & \includegraphics[width=0.15\linewidth,trim={51cm 13.5cm 0.5cm 2.9cm}, clip]{figures/case_drug.pdf}\\
        & QED: 0.79 & QED: 0.79 & QED: 0.77 & QED: 0.77 & QED: 0.63\\
        \rotatebox{90}{3M-Diffusion}&  \includegraphics[width=0.15\linewidth,trim={8cm 2cm 44cm 13cm}, clip]{figures/case_drug.pdf} &  \includegraphics[width=0.15\linewidth,trim={19cm 2cm 36cm 13cm}, clip]{figures/case_drug.pdf} &  \includegraphics[width=0.15\linewidth,trim={30cm 2cm 24cm 13cm}, clip]{figures/case_drug.pdf} &  \includegraphics[width=0.15\linewidth,trim={40cm 2cm 13cm 13cm}, clip]{figures/case_drug.pdf} &  \includegraphics[width=0.15\linewidth,trim={50cm 2cm 1cm 13cm}, clip]{figures/case_drug.pdf}\\
        & QED: 0.89 & QED: 0.83 & QED: 0.80 & QED: 0.78 & QED: 0.76\\
          
    \end{tabular}
    }
    \caption{Comparative analysis of generated molecules from Mol-CADiff, 3M-Diffusion, and MolT5-large models under drug likeliness condition (best viewed when zoomed in). The top 5 molecules are selected based on desired properties, with drug likeliness measured by QED (higher values indicate better drug likeliness). Our model shows the best performance.}
    \label{tab:case_drug}
\end{table*}

\begin{table*}[!ht]
    \centering
    \resizebox{\textwidth}{!}{
    \begin{tabular}{lccccc}
        & \multicolumn{5}{c}{\textbf{Condition:} This molecule is not like a drug.}\\
         \rotatebox{90}{Mol-CADiff}&  \includegraphics[width=0.15\linewidth,trim={0cm 0.8cm 0cm 1.1cm}, clip]{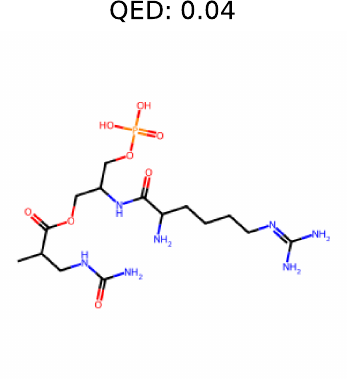}&  \includegraphics[width=0.15\linewidth,trim={0cm 0.8cm 0cm 1.1cm}, clip]{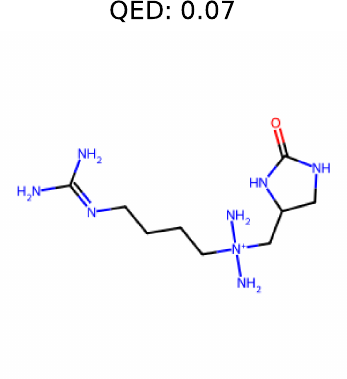}&  \includegraphics[width=0.15\linewidth,trim={0cm 1cm 0cm 1.1cm}, clip]{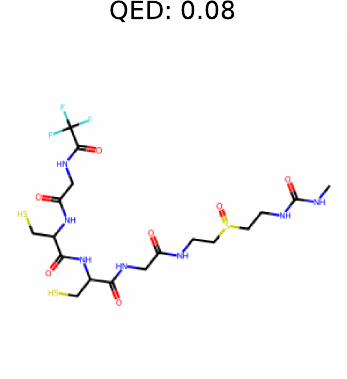}&  \includegraphics[width=0.15\linewidth,trim={0cm 0.8cm 0cm 1.1cm}, clip]{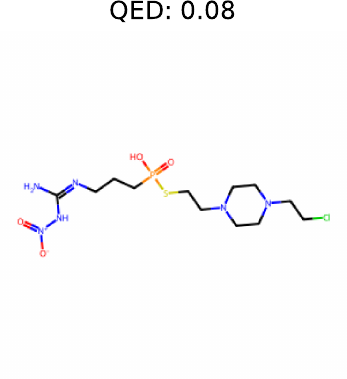}&  \includegraphics[width=0.15\linewidth,trim={0cm 0.8cm 0cm 1.1cm}, clip]{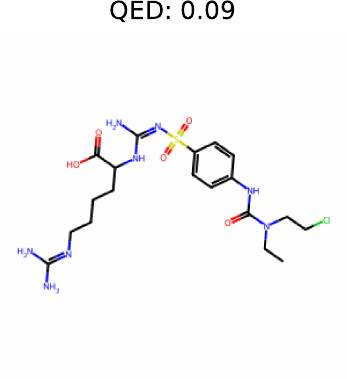}\\
         & QED: 0.04 & QED: 0.07 & QED: 0.08 & QED: 0.08 & QED: 0.09\\
        \rotatebox{90}{MolT5-large}&  \includegraphics[width=0.15\linewidth,trim={9cm 14cm 43cm 2.9cm}, clip]{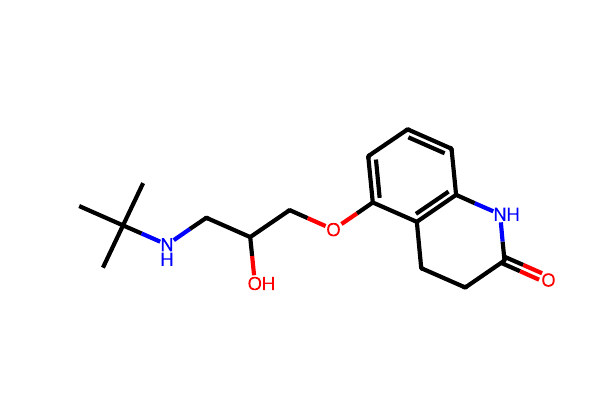} & \includegraphics[width=0.15\linewidth,trim={19cm 14cm 33cm 2.9cm}, clip]{figures/case_indrug.pdf} & \includegraphics[width=0.15\linewidth,trim={29cm 14cm 23cm 2.9cm}, clip]{figures/case_indrug.pdf} & \includegraphics[width=0.15\linewidth,trim={41cm 14cm 11cm 2.9cm}, clip]{figures/case_indrug.pdf} & \includegraphics[width=0.15\linewidth,trim={51cm 14cm 1cm 2.9cm}, clip]{figures/case_indrug.pdf}\\
        & QED: 0.77 & QED: 0.77 & QED: 0.89 & QED: 0.89 & QED: 0.89\\
        \rotatebox{90}{3M-Diffusion}&  \includegraphics[width=0.15\linewidth,trim={8cm 2cm 44cm 14cm}, clip]{figures/case_indrug.pdf} &  \includegraphics[width=0.15\linewidth,trim={19cm 2cm 34cm 14cm}, clip]{figures/case_indrug.pdf} &  \includegraphics[width=0.15\linewidth,trim={29cm 2cm 23cm 14cm}, clip]{figures/case_indrug.pdf} &  \includegraphics[width=0.15\linewidth,trim={40cm 2cm 11cm 14cm}, clip]{figures/case_indrug.pdf} &  \includegraphics[width=0.15\linewidth,trim={51cm 2cm 0.5cm 14cm}, clip]{figures/case_indrug.pdf}\\
        & QED: 0.05 & QED: 0.05 & QED: 0.11 & QED: 0.11 & QED: 0.13\\
          
    \end{tabular}
    }
    \caption{Comparative analysis of generated molecules from Mol-CADiff, 3M-Diffusion, and MolT5-large models under drug unlikeliness condition (best viewed when zoomed in). The top 5 molecules are selected based on desired properties, with drug likeliness measured by QED (lower values indicate better drug unlikeliness). Our model shows the best performance.}
    \label{tab:case_undrug}
\end{table*}

\begin{table*}[!ht]
    \centering
    \resizebox{\textwidth}{!}{
    \begin{tabular}{lccccc}
        & \multicolumn{5}{c}{\textbf{Condition:} The molecule is an anthocyanidin cation.}\\
         \rotatebox{90}{Mol-CADiff}&  \includegraphics[width=0.15\linewidth,trim={8cm 14cm 43cm 2.9cm}, clip]{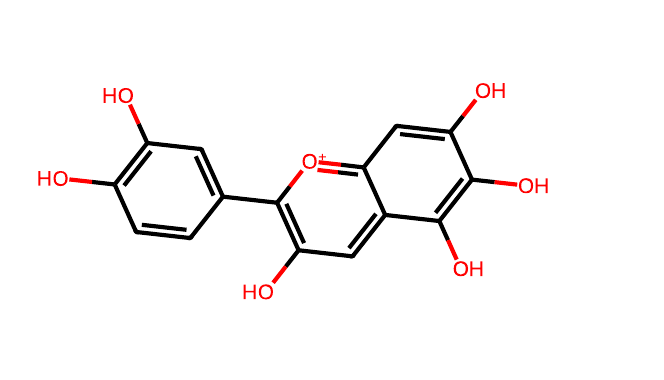}&  \includegraphics[width=0.15\linewidth,trim={0cm 0.8cm 0cm 1.1cm}, clip]{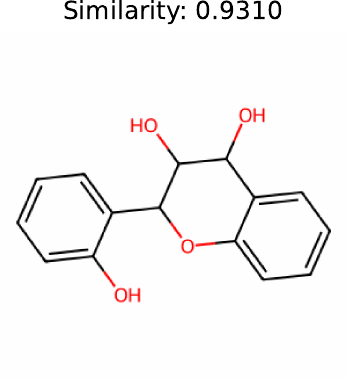}&  \includegraphics[width=0.15\linewidth,trim={0cm 1cm 0cm 1.1cm}, clip]{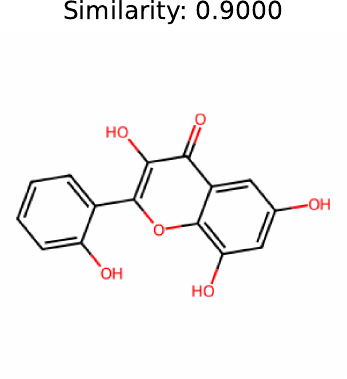}&  \includegraphics[width=0.15\linewidth,trim={0cm 0.8cm 0cm 1.1cm}, clip]{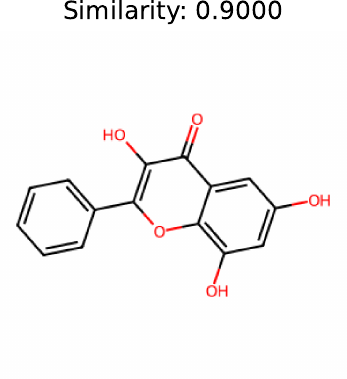}&  \includegraphics[width=0.15\linewidth,trim={0cm 0.8cm 0cm 1.1cm}, clip]{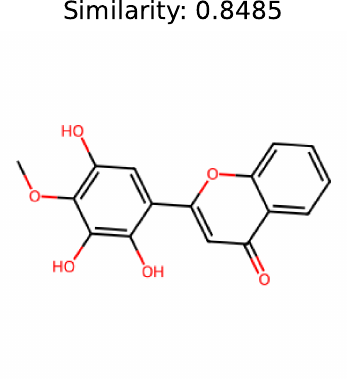}\\
         & Reference & Sim: 0.93& Sim: 0.90 & Sim: 0.90 & Sim: 0.85\\
        \rotatebox{90}{MolT5-large}&  \includegraphics[width=0.15\linewidth,trim={8cm 14cm 43cm 2.9cm}, clip]{figures/case_an.pdf} & \includegraphics[width=0.15\linewidth,trim={18cm 14cm 33cm 2.5cm}, clip]{figures/case_an.pdf} & \includegraphics[width=0.15\linewidth,trim={28cm 14cm 23cm 2.5cm}, clip]{figures/case_an.pdf} & \includegraphics[width=0.15\linewidth,trim={40cm 14cm 11cm 2.5cm}, clip]{figures/case_an.pdf} & \includegraphics[width=0.15\linewidth,trim={52cm 14cm 1cm 2.5cm}, clip]{figures/case_an.pdf}\\
        & Reference &  Sim: 0.74 & Sim: 0.74 & Sim: 0.74 & Sim: 0.73\\
        \rotatebox{90}{3M-Diffusion}&   \includegraphics[width=0.15\linewidth,trim={8cm 14cm 43cm 2.9cm}, clip]{figures/case_an.pdf} & \includegraphics[width=0.15\linewidth,trim={18cm 2.5cm 33cm 14cm}, clip]{figures/case_an.pdf} & \includegraphics[width=0.15\linewidth,trim={28cm 2.5cm 23cm 14cm}, clip]{figures/case_an.pdf} & \includegraphics[width=0.15\linewidth,trim={40cm 2.5cm 11cm 14cm}, clip]{figures/case_an.pdf} & \includegraphics[width=0.15\linewidth,trim={52cm 2.5cm 1cm 14cm}, clip]{figures/case_an.pdf}\\
        & Reference &  Sim: 0.87 & Sim: 0.85 & Sim: 0.83 & Sim: 0.81\\
          
    \end{tabular}
    }
    \caption{Comparative analysis of generated molecules from Mol-CADiff, 3M-Diffusion, and MolT5-large models under anthocyanidin cation condition (best viewed when zoomed in). Reference is selected as a ground truth. A higher similarity (Sim.) indicates better alignment with the reference. Our model shows the best performance.}
    \label{tab:case_an}
\end{table*}

\end{document}